# Generative adversarial networks with Monte Carlo dropout uncertainty quantification for structural health monitoring: Fibre Bragg grating data augmentation comparison in Hawk T1A aircraft

Alvaro Peiro, Marios Impraimakis *

*Department of Mechanical Engineering, University of Bath, Bath, BA2 7AY, UK*

A B S T R A C T

The signal generation capabilities of a novel data augmentation framework are examined in aerospace structures. First, the framework learns class-conditional temporal and spectral characteristics of the measured structural responses. It uses conditional generative adversarial network training and Monte Carlo dropout uncertainty quantification as a Bayesian approximation. This step enables stochastic signal synthesis and uncertainty quantification. Second, the generated responses are rigorously compared against experimental measurements across time, frequency, uncertainty, variability, and feature domains to assess physical realism, statistical consistency, and distributional fidelity. Finally, the synthetic data are evaluated to determine whether they should be incorporated into a supervised damage classification workflow. The framework is validated on the full-scale benchmark dataset from a BAE Systems Hawk T1A aircraft tested at the University of Sheffield's Structural Dynamics Laboratory for Verification and Validation. The test involves acceleration, fibre Bragg grating strain, and nonlinear dynamic response measurements. The results demonstrate that the model accurately captures dominant structural dynamics, it guides on signal use for classification improvement based on uncertainty characteristics, and it provides meaningful uncertainty estimates linked to signal complexity and excitation characteristics. Importantly, the framework offers a realistic generative modelling plan as a robust digital twin component for data-driven structural health monitoring, supporting resilient decision-making for condition assessment in safety-critical aerospace systems.

## 1. Introduction

The reliable detection of structural damage in complex engineering systems remains a fundamental challenge as it is influenced by a wide range of operational and environmental factors [1]. In practice, engineering structures are designed to operate in healthy conditions [2]. The datasets, therefore, are heavily biased towards undamaged states [3]. Obtaining labelled damage data is costly, complex, and constrained by practical and experimental limitations [4,5].

In the aerospace sector [6], structural health monitoring is particularly critical due to stringent safety requirements, complex loading environments, and exposure to variable operational conditions [7–9]. While these factors increase the need for structural health monitoring, they also amplify challenges related to reliability, validation, and limited availability of representative damage data [10]. Importantly, the structural health monitoring methodologies are commonly categorised into physics-driven and data-driven approaches, with hybrid methods emerging to combine elements of both [11–13]. This distinction highlights a key trade-off between the two approaches: model-based methods offer interpretability but struggle to represent complex structures and environmental variability. Data-driven methods, on the other side, capture complex patterns but depend heavily on representative data [5]. Hybrid approaches aim to balance these strengths, combining physical insight with data-driven flexibility [14]. Furthermore, vibration-based approaches are widely adopted due to their robustness and simplicity. However, their effectiveness is constrained by challenges in accurately identifying damage characteristics, along with the need for extensive data processing and real-world validation [15]. While their ability to capture global structural behaviour makes them well suited to large-scale systems [16], it reduces the sensitivity to localised or early-stage damage [17].

To address data scarcity, generative modelling approaches have been explored to synthesise additional training data [18–22]. Generative adversarial networks, GANs, by Goodfellow et al. [23] enable the learning of underlying data distributions and the generation of synthetic samples. Conditional generative adversarial networks by Mirza et al. [24]

* Corresponding author.
*E-mail address:* mi595@bath.ac.uk (M. Impraimakis).

allow generation of labelled data corresponding to different structural states. Recent work by Luleci et al. [25] has focused on generative modelling techniques, which aim to learn the underlying data distribution and generate realistic synthetic samples [26]. Recurrent neural networks, particularly the long short-term memory networks, can model temporal dependencies and capture sequential patterns in vibration data [27,28]. However, as they are typically trained to minimise reconstruction or prediction error, they tend to reproduce observed data rather than learn the full underlying distribution, limiting their ability to generate truly novel samples representative of unseen damage conditions [29]. In addition, their sequence-based formulation restricts their ability to model the joint behaviour of multi-sensor responses, which is critical in large-scale structural health monitoring systems [30]. Variational autoencoders provide a probabilistic framework by learning a latent representation of the data [31]. While they enable controlled sample generation, they often produce overly smooth signals that lack high-frequency detail. This reduces their effectiveness for detecting subtle or localised damage where important features are embedded in higher-frequency components [17,32]. Generative adversarial networks offer a powerful alternative for generating high-fidelity data through adversarial training [23]. In structural health monitoring, they have demonstrated the ability to generate realistic vibration signals and capture complex structural behaviour [33]. Conditional generative adversarial networks are also well suited to modelling complex, high-dimensional time-series data and, unlike reconstruction-based models, optimise generation quality through adversarial learning [34].

To this end, this work investigates the use of uncertainty-aware conditional generative adversarial networks for augmenting structural vibration data in structural health monitoring. It is validated on a recently developed full-scale benchmark dataset based on the Hawk T1A aircraft [35–39], which is known for its use by the Royal Air Force Red Arrows display team. The Hawk T1A aircraft dataset by Wilson et al. [35] provides high-dimensional and complex structural response data, characterised by nonlinear behaviour and significant variability, which are difficult to model accurately. The generated signals are evaluated in terms of their physical realism and statistical consistency across time, frequency, uncertainty, variability, and feature domains, and are further assessed through their impact on the performance of a structural state classification model by providing a framework that judges on whether the generated signals should be utilised.

The work is organised as follows: Section 2 describes the experimental design of the BAE Systems Hawk T1A aircraft. Section 3 presents the uncertainty-aware conditional generative adversarial network framework, including the model architecture, the training strategy, and the uncertainty quantification. Section 4 details the evaluation metrics used to assess signal realism and statistical consistency. Section 5 applies the proposed method to fibre Bragg grating strain, acceleration, and nonlinear dynamic response measurements under multiple excitation and damage scenarios. Section 6 discusses the findings, limitations, and implications of the proposed approach. Finally, Section 7 concludes the work.

## 2. The experimental design of the BAE Systems Hawk T1A aircraft

A full-scale structural vibration test dataset is employed by a decommissioned BAE Systems Hawk T1A aircraft. The experimental measurements were collected at the University of Sheffield's Laboratory for Verification and Validation [35]. Representative frequency response functions are also shown by Wilson et al. [35]. In total, it comprises 216 distinct test configurations, each repeated approximately ten times to assess measurement repeatability. Each test includes approximately 139 synchronised measurement channels, including accelerometers, fibre Bragg grating FBG strain gauges, force transducers, and environmental sensors. The structural variation is introduced through baseline healthy, pseudo-damage, and physically modified conditions. The pseudo-damage is implemented via discrete mass additions at selected locations, applied in both single-site and multi-site configurations. A physical modification is additionally introduced through panel removal along the port wing.

**Table 1**
Selected subset of experiments, sensor locations, and corresponding measurement and damage conditions used in this study.

| Experiment | Sensor | Measurement | Damage Location | Excitation Location |
|---|---|---|---|---|
| ORPM02-S | PW_FB2 | Strain | SW_TLE | Starboard wing |
| ORPM02-A | PW_ULE3 | Acceleration | SW_TLE | Starboard wing |
| WN03-S | PW_FB2 | Strain | PW1 | Global |
| WN03-A | PW_ULE3 | Acceleration | PW1 | Global |

Fig. 2 provides representative frequency response functions of the healthy structure for the configurations investigated in this work. The dominant resonances are concentrated within the primary excitation bandwidth of 5–256 Hz, providing additional characterisation of the structural dynamics and experimental setup used throughout the study.

The excitation is applied using multiple electrodynamic shakers positioned to excite the wings, stabilisers, and rudder, enabling both single-input multiple-output and multiple-input multiple-output configurations. A range of broadband excitation signals is employed, including white noise, pink noise, and odd random-phase multisine ORPM inputs. ORPM excitation is specifically used to probe non-linear system behaviour, as its structured frequency content enables identification of harmonic distortions and energy transfer to unexcited frequencies. All tests are conducted over a frequency range of 5–256 Hz, with measurements recorded at a sampling rate of 2048 Hz.

In this work, two experimental configurations are selected: WN03 broadband white noise and ORPM02 odd random-phase multisine, following the naming convention in [35]. In this dataset, 'S' and 'A' denote the strain and acceleration measurements at the selected sensor location, as summarised in Table 1, with the full sensor arrangement and damage locations shown in Fig. 1. The selected datasets are chosen to provide a representative combination of excitation types, sensing modalities, and damage scenarios seen in Table 2.

Specifically, the white noise excitation provides a stochastic broadband response suitable for assessing the ability of the model to reproduce complex vibration behaviour, while the ORPM excitation is commonly employed for analysing nonlinear structural dynamics due to its controlled spectral characteristics. In the present study, one FBG sensor (PW_FB2) and one accelerometer (PW_ULE3) are selected to provide measurements of local strain and global structural vibration behaviour. The investigated damage scenarios include both pseudo-damage introduced through discrete mass additions at the starboard wing leading edge (SW_TLE) and physical damage introduced through panel removal at location PW1 on the port wing. These scenarios are selected as they produce distinct structural response characteristics while representing different classes of structural modification.

Furthermore, each dataset consists of continuous time series recordings of 163,840 samples (80 s), with ten repeated tests for each of the healthy and damaged conditions. To prevent data leakage between training and testing datasets, the train-test split is performed at the recording level prior to window segmentation, as summarised in Table 3. The recordings are subsequently segmented into fixed-length non-overlapping windows, yielding 788 healthy and 787 damaged samples. Each window is truncated to 1201 time steps. The data are then mean-centred and normalised using the global standard deviation. Specifically, each vibration measurement is represented as a discrete time series:

$$x_i[n], \quad i = 1, \dots, N, \ n = 1, \dots, T \tag{1}$$

where, $T = 1201$ is the number of retained time steps. Mean centering is applied to each signal as:

$$\tilde{x}_i[n] = x_i[n] - \frac{1}{T}\sum_{n=1}^{T} x_i[n] \tag{2}$$

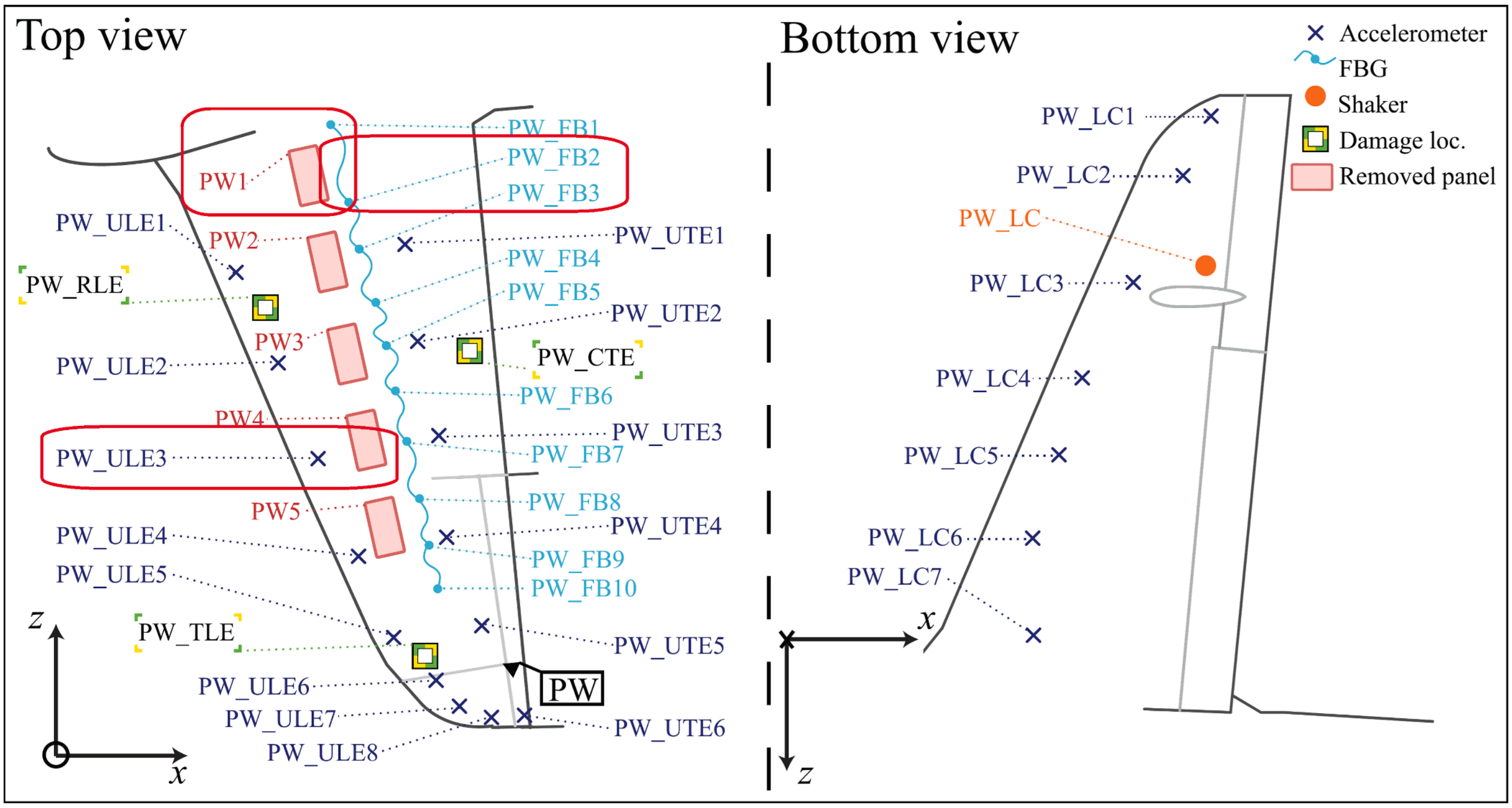

**Fig. 1.** Port wing sensor layout showing top and bottom views of the experimental setup, including accelerometer and FBG sensor locations, excitation point, and damage locations [35]. The red boxes highlight the specific sensor locations (PW_FB2 and PW_ULE3) and damage locations (PW1 and SW_TLE) used in the present study. (For interpretation of the references to colour in this figure legend, the reader is referred to the web version of this article.)

**Table 2**
Detailed information for the selected experimental configurations.

| Dataset | Excitation | Sensor | Structural Damage |
|---|---|---|---|
| ORPM02-S | ORPM excitation over 5–256 Hz for nonlinear structural behaviour. | PW_FB2 on the port wing. | Mass addition at SW_TLE. |
| ORPM02-A | ORPM over 5–256 Hz for nonlinear structural behaviour. | PW_ULE3 on the port wing. | Mass addition at SW_TLE. |
| WN03-S | WN over 5–256 Hz for stochastic structural responses. | PW_FB2 on the port wing. | Panel removal at PW1. |
| WN03-A | WN over 5–256 Hz for stochastic structural responses. | PW_ULE3 on the port wing | Panel removal at PW1. |

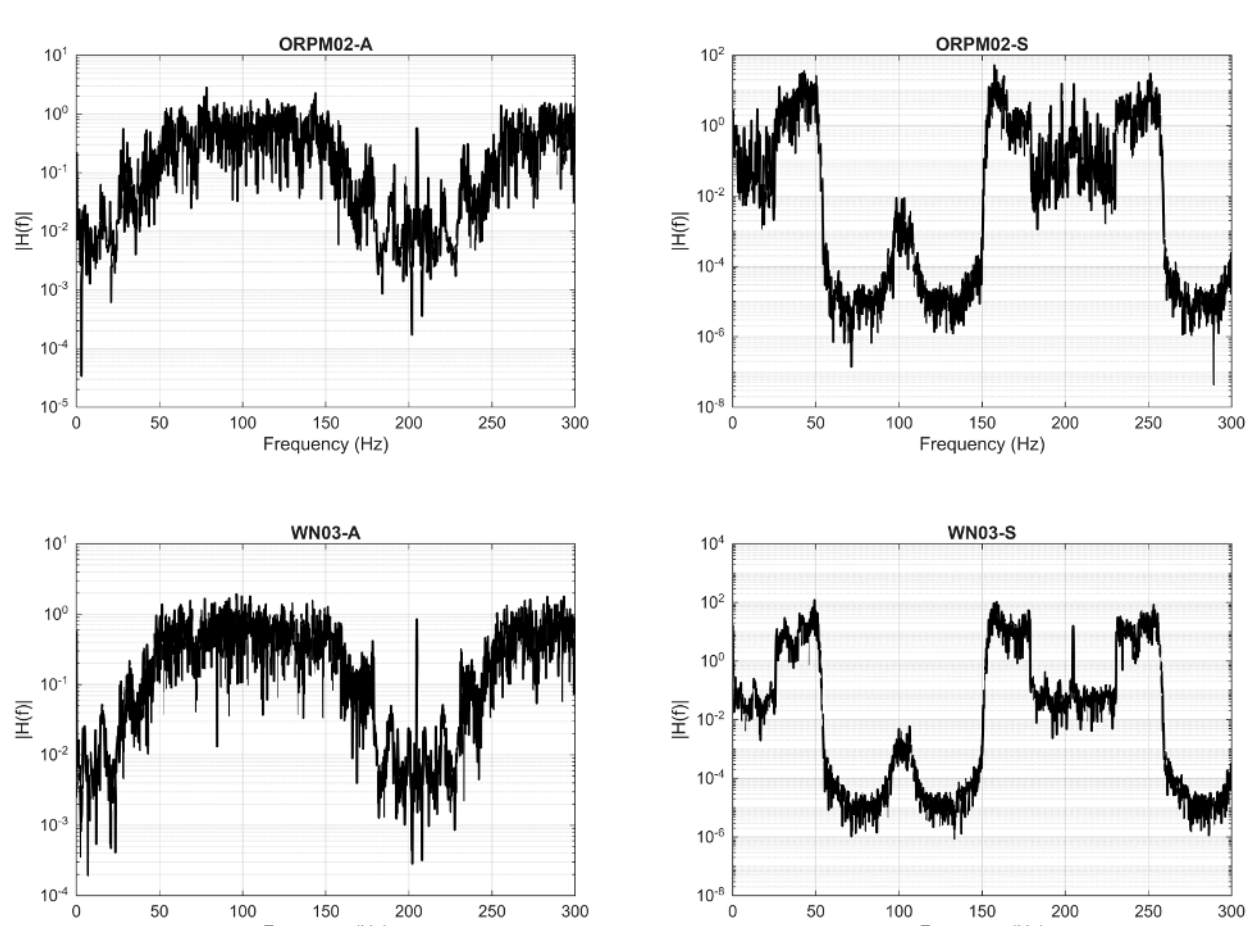

**Fig. 2.** Representative healthy-state frequency response functions of the Hawk T1A aircraft obtained from measured force and response data. Wilson et al. [35] discusses the complexity of the Hawk structure and computes FRFs from cross-spectral estimates with processing and averaging.

while, the centred signals are normalised using the global standard deviation as:

$$x_i^{\text{norm}}[n] = \frac{\tilde{x}_i[n]}{\sigma_{\text{global}}} \quad (3)$$

**Table 3**
Data partitioning procedure.

| Processing step | Description |
|---|---|
| Recording selection | Healthy and damaged measurements are separated into training (70%) and testing (30%) subsets. |
| Window segmentation | Each recording is segmented into fixed-length non-overlapping windows. |
| Leakage prevention | Windows originating from the same experimental recording are not present in both training and testing datasets. |
| Evaluation | Classification performance is evaluated using windows derived from unseen test recordings. |

with

$$\sigma_{\text{global}} = \sqrt{\frac{1}{NT}\sum_{i=1}^{N}\sum_{n=1}^{T}\tilde{x}_i^2[n]}. \quad (4)$$

## 3. Uncertainty-aware generative adversarial network modeling

The examined framework learns and reproduces class-dependent vibration characteristics under varying excitation and damage conditions. The generative adversarial network extends a conditional GAN architecture [40]. The model consists of a generator and discriminator trained adversarially, enabling the model to learn the joint distribution of vibration signals and structural states. The generator $G$ maps latent noise and class labels to synthetic signals, while the discriminator evaluates their consistency, forcing the generator to produce physically and statistically plausible outputs, written as:

$$\hat{x} = G(z, y) \tag{5}$$

where, $z \sim \mathcal{N}(0, 1)$ and $y$ denotes the structural condition [23,24]. The latent vector is projected to form an initial feature representation, while the label is embedded to incorporate structural condition information. The resulting representations are concatenated and passed through a series of transposed convolution layers [41], which apply learned filters to progressively refine the feature representations and produce a coherent vibration signal. The architecture is shown in Fig. 3, with layer dimensions reported as length × width × channels, where the number of channels corresponds to the number of filters in the preceding transposed convolution layer.

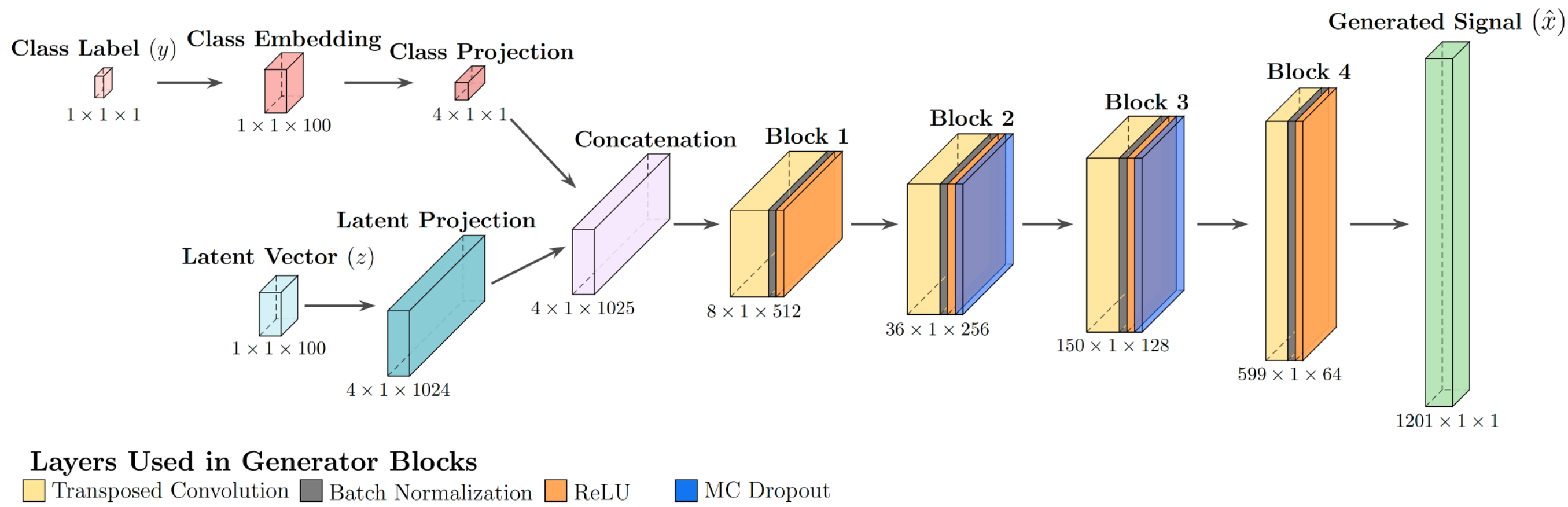

**Fig. 3.** Examined generator architecture.

Batch normalisation and Rectified Linear Unit, ReLU, activations are used throughout the network to stabilise training and enable the learning of non-linear feature representations [42,43]. The Monte Carlo dropout layer is then incorporated into the base architecture [40] as a modification to reduce overfitting and provide approximate uncertainty estimates [44,45]. By retaining dropout during inference, the generator produces stochastic outputs for identical inputs, allowing uncertainty in generated signals to be assessed as:

$$\tilde{h}^{(l)} = h^{(l)} \odot \mathbf{m}^{(l)} \quad \mathbf{m}^{(l)} \sim \text{Bernoulli}(1-p) \tag{6}$$

For a fixed latent vector $z$ and class label $y$, $K$ stochastic forward passes are performed using different dropout mask realisations as:

$$\hat{x}^{(k)} = G(z, y; \mathbf{m}^{(k)}), \quad k = 1, \dots, K \tag{7}$$

$$\mu[n] = \frac{1}{K} \sum_{k=1}^{K} \hat{x}^{(k)}[n] \tag{8}$$

$$\sigma^2[n] = \frac{1}{K-1} \sum_{k=1}^{K} \left(\hat{x}^{(k)}[n] - \mu[n]\right)^2 \tag{9}$$

$$U = \frac{1}{T} \sum_{n=1}^{T} \sigma[n] \tag{10}$$

$$U_{\text{norm}} = \frac{U}{\frac{1}{T} \sum_{n=1}^{T} |\mu[n]|} \tag{11}$$

where, $h^{(l)}$ denotes the activation at layer $l$, $\mathbf{m}^{(l)}$ is a dropout mask drawn from a Bernoulli distribution with retention probability $1-p$, and $\odot$ represents element-wise multiplication. The generator $G(\cdot)$ maps the latent vector $z$ and class label $y$ to a synthetic signal, with $\mathbf{m}^{(k)}$ indicating the dropout realisation used during the $k$th stochastic forward pass. The $K$ generated signals $\hat{x}^{(k)}[n]$ are used to compute the sample mean $\mu[n]$ and variance $\sigma^2[n]$ at each time step $n$. A global uncertainty measure $U$ is obtained by averaging the pointwise standard deviation over the signal length $T$, while $U_{\text{norm}}$ normalises this quantity by the mean absolute signal amplitude to enable comparison across datasets with different signal magnitudes. The uncertainty estimation procedure is summarised in Table 4. A dropout rate of $p = .2$ is also applied within intermediate layers to provide moderate regularisation without destabilising adversarial training [46]. The architecture shown in Table 5 is structured to capture the range of vibration responses observed in the Hawk dataset. The use of different kernel sizes allows the network to capture patterns over different portions of the signal, while the strides control how the signal is progressively expanded during generation.

**Table 4**
Pseudo-algorithm for Monte Carlo dropout uncertainty quantification.

**Algorithm 1: Monte Carlo dropout uncertainty quantification.**

**Input:** Trained generator $G$, latent vector $z$, structural condition label $y$, dropout probability $p = .2$, number of stochastic forward passes $K = 100$
**Output:** Mean response $\mu[n]$, uncertainty estimate $U$, normalised uncertainty $U_{\text{norm}}$
1: Retain dropout layers during inference
2: Fix latent vector $z$ and structural condition label $y$
3: For $k = 1, \dots, K$ do
  3.1: Sample dropout mask $\mathbf{m}^{(k)} \sim \text{Bernoulli}(1-p)$
  3.2: Generate stochastic response $\hat{x}^{(k)} = G(z, y; \mathbf{m}^{(k)})$
4: End for
5: Compute ensemble mean $\mu[n] = \frac{1}{K} \sum_{k=1}^{K} \hat{x}^{(k)}[n]$.
6: Compute ensemble variance $\sigma^2[n] = \frac{1}{K-1} \sum_{k=1}^{K} \left(\hat{x}^{(k)}[n] - \mu[n]\right)^2$
7: Compute global uncertainty $U = \frac{1}{T} \sum_{n=1}^{T} \sigma[n]$
8: Compute normalised uncertainty $U_{\text{norm}} = \frac{U}{\frac{1}{T} \sum_{n=1}^{T} |\mu[n]|}$
9: Use $U_{\text{norm}}$ to compare uncertainty across datasets and structural states

On the other hand, the discriminator network is responsible for distinguishing between real vibration signals from the dataset and synthetic signals generated by the generator. It learns a mapping $D(x, y)$, where $x$ represents a vibration signal and $y$ is the corresponding class label, following the conditional generative adversarial framework [24]. The discriminator outputs a scalar score indicating whether the input signal-label pair is real or generated. This score is passed through a sigmoid activation function to obtain a probability in the range $[0, 1]$, which is used in the adversarial loss function to train both the discriminator and generator [23]. The architecture of the discriminator is illustrated in Fig. 4.

Here, the conditional information is incorporated by embedding the class label and reshaping it to match the signal dimensions, before concatenation with the input signal along the channel dimension. The use of convolutional layers is well-suited to vibration signals, as it allows local patterns within the time-series to be detected and related to the signal behaviour [41]. The selection of kernel sizes and strides in Table 6 supports this process. Larger kernels in earlier layers allow broader sections of the signal to be considered, while smaller kernels in deeper layers focus on more detailed signal features. Strided convolutions progressively reduce the 1201-sample input, enabling the network to form a compact representation while retaining features associated with the broadband and complex structural response of the Hawk dataset [35], which are required to distinguish between real and generated signals. Leaky ReLU activations are then applied after each convolutional layer to introduce non-linear transformations while preventing neurons from becoming inactive for negative inputs, improving training stability [47].

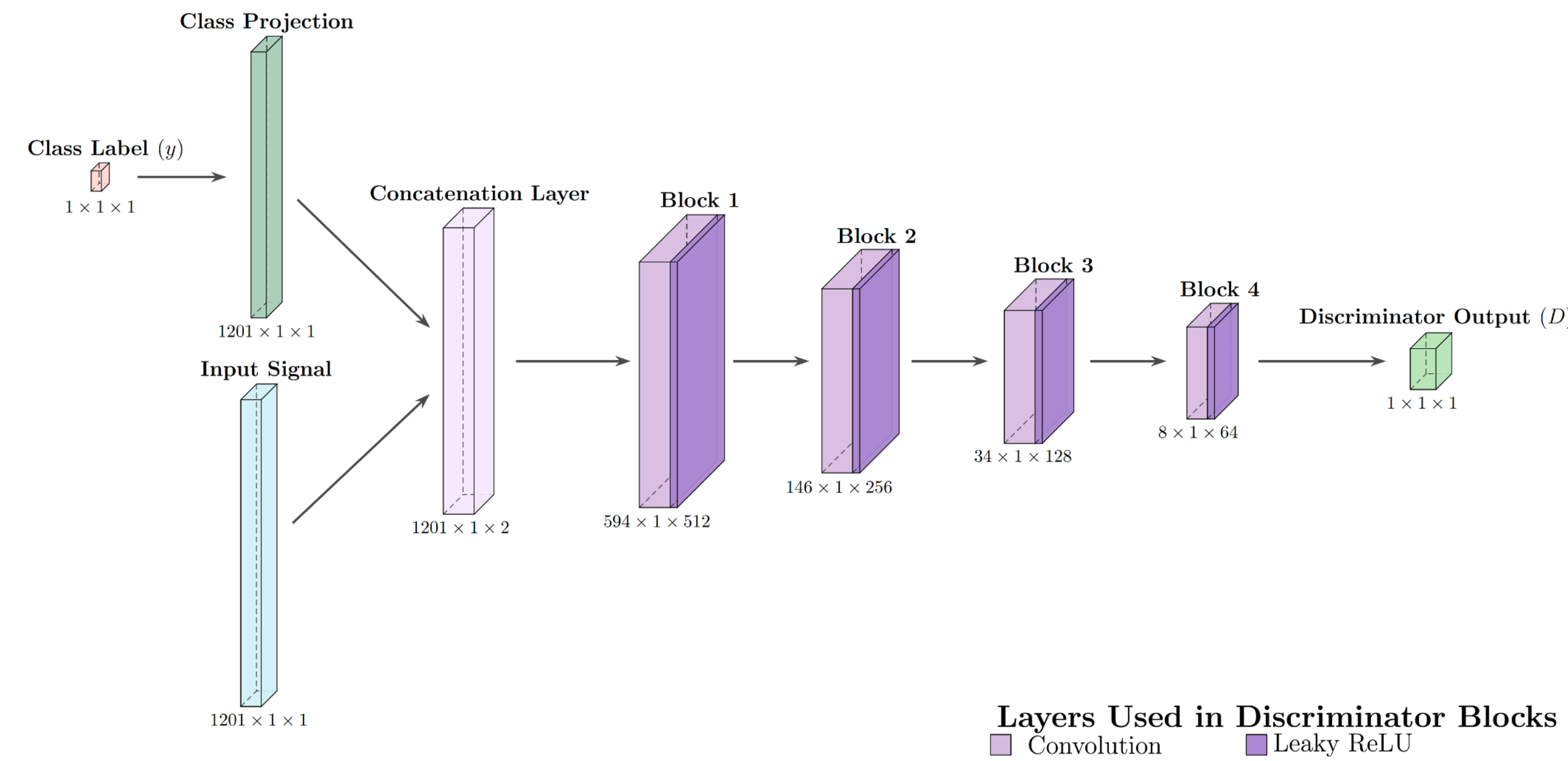

**Fig. 4.** Examined discriminator architecture.

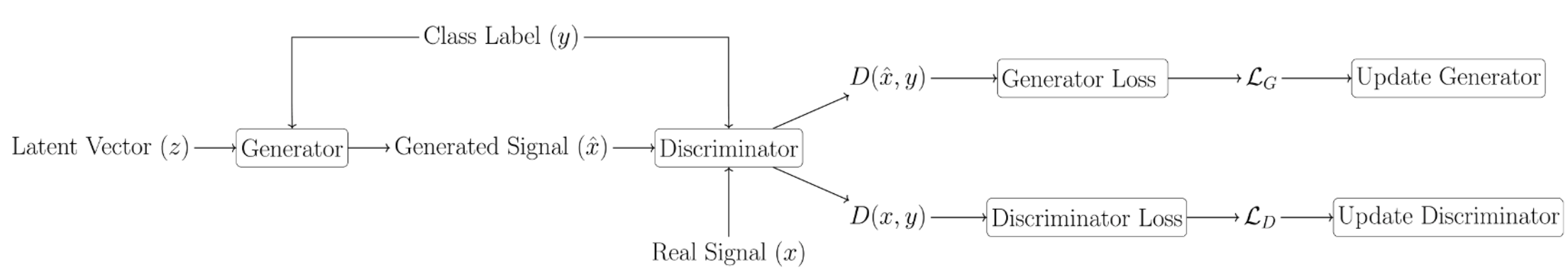

**Fig. 5.** Adversarial training workflow for the conditional generative adversarial network.

**Table 5**
Generator network architecture used for conditional vibration signal synthesis.

| Layer | Type | Kernel | Stride | Output Size |
|---|---|---|---|---|
| in | Input (latent vector) | – | – | $1 \times 1 \times 100$ |
| proj | Project and reshape | – | – | $4 \times 1 \times 1024$ |
| labels | Label input | – | – | $1 \times 1 \times 1$ |
| emb | Label embedding reshape | – | – | $4 \times 1 \times 1$ |
| cat | Concatenation | – | – | $4 \times 1 \times 1025$ |
| tconv1 | Transposed convolution | $5 \times 1$ | 1 | $8 \times 1 \times 512$ |
| bn1 | Batch normalisation | – | – | $8 \times 1 \times 512$ |
| relu1 | ReLU activation | – | – | $8 \times 1 \times 512$ |
| tconv2 | Transposed convolution | $10 \times 1$ | 4 | $36 \times 1 \times 256$ |
| bn2 | Batch normalisation | – | – | $36 \times 1 \times 256$ |
| relu2 | ReLU (Rectified Linear Unit) activation | – | – | $36 \times 1 \times 256$ |
| mcDropG1 | MC Dropout ($p = .2$) | - | – | $36 \times 1 \times 256$ |
| tconv3 | Transposed convolution | $12 \times 1$ | 4 | $150 \times 1 \times 128$ |
| bn3 | Batch normalisation | – | – | $150 \times 1 \times 128$ |
| relu3 | ReLU activation | – | – | $150 \times 1 \times 128$ |
| mcDropG2 | MC Dropout ($p = .2$) | – | – | $150 \times 1 \times 128$ |
| tconv4 | Transposed convolution | $5 \times 1$ | 4 | $599 \times 1 \times 64$ |
| bn4 | Batch normalisation | – | – | $599 \times 1 \times 64$ |
| relu4 | ReLU activation | – | – | $599 \times 1 \times 64$ |
| tconv5 | Transposed convolution | $7 \times 1$ | 2 | $1201 \times 1 \times 1$ |

Finally, the generator and discriminator networks are trained through an adversarial optimisation process, in which both models are updated iteratively within a competitive framework [23]. At each itera-tion, the generator produces synthetic vibration signals conditioned on structural state, while the discriminator evaluates their consistency with real signal-label pairs. This iterative interaction, illustrated in Fig. 5, enables the model to learn the underlying data distribution.

**Table 6**
Discriminator network architecture used to evaluate real and generated vibration signals.

| Layer | Type | Kernel | Stride | Output Size |
|---|---|---|---|---|
| in | Input signal | – | – | $1201 \times 1 \times 1$ |
| labels | Label input | – | – | $1 \times 1 \times 1$ |
| emb | Label embedding reshape | – | – | $1201 \times 1 \times 1$ |
| cat | Concatenation | – | – | $1201 \times 1 \times 2$ |
| conv1 | Convolution | $17 \times 1$ | 2 | $594 \times 1 \times 512$ |
| lrelu1 | Leaky ReLU | – | – | $594 \times 1 \times 512$ |
| conv2 | Convolution | $16 \times 1$ | 4 | $146 \times 1 \times 256$ |
| lrelu2 | Leaky ReLU | – | – | $146 \times 1 \times 256$ |
| conv3 | Convolution | $16 \times 1$ | 4 | $34 \times 1 \times 128$ |
| lrelu3 | Leaky ReLU | – | – | $34 \times 1 \times 128$ |
| conv4 | Convolution | $8 \times 1$ | 4 | $8 \times 1 \times 64$ |
| lrelu4 | Leaky ReLU | – | – | $8 \times 1 \times 64$ |
| conv5 | Convolution | $8 \times 1$ | 1 | $1 \times 1 \times 1$ |

**Table 7**
Training hyperparameters.

| Dataset | Batch Size | Generator LR | Discriminator LR | Epochs |
|---|---|---|---|---|
| ORPM02-S | 256 | $3 \times 10^{-4}$ | $2 \times 10^{-4}$ | 1500 |
| ORPM02-A | 256 | $3 \times 10^{-4}$ | $3 \times 10^{-4}$ | 1000 |
| WN03-S | 256 | $3 \times 10^{-4}$ | $2 \times 10^{-4}$ | 1500 |
| WN03-A | 256 | $1.5 \times 10^{-4}$ | $2 \times 10^{-4}$ | 2000 |

The adversarial objective is implemented using a binary cross-entropy loss as:

$$L_D = -\mathbb{E}_{x\sim p_{\text{data}}}\left[\log D(x,y)\right] - \mathbb{E}_{z\sim p_z}\left[\log\left(1 - D(G(z,y),y)\right)\right] \tag{12}$$

$$L_G = -\mathbb{E}_{z\sim p_z}\left[\log D(G(z,y),y)\right] \tag{13}$$

$$\min_G \max_D \ \mathbb{E}_{x\sim p_{\text{data}}}[\log D(x,y)] + \mathbb{E}_{z\sim p_z}[\log(1 - D(G(z,y),y))] \tag{14}$$

The model parameters are optimised using the Adam optimiser [48], due to its stable convergence behaviour in adversarial training. Momentum parameters of $\beta_1 = 0.5$ and $\beta_2 = 0.999$ are adopted to reduce oscillations while incorporating the influence of previous gradient updates. The selected hyperparameters are summarised in Table 7.

## 4. Evaluation metrics

The training behaviour is assessed through generator and discriminator scores, defined as the average discriminator probabilities for generated and real samples, written as:

$$S_G = \mathbb{E}_{z\sim p_z}\left[D(G(z,y),y)\right] \tag{15}$$

$$S_D = \mathbb{E}_{x\sim p_{\text{data}}}\left[D(x,y)\right] \tag{16}$$

As shown in Fig. 6, early training is characterised by strong discriminator performance, with generated signals easily distinguished from real data. As training progresses, the generator improves, making discrimination increasingly difficult. At later stages, both scores stabilise with bounded variation, suggesting a balanced adversarial state in which neither network consistently dominates. It should be noted that adversarial training does not inherently guarantee convergence and may be susceptible to discriminator domination, generator domination, oscillatory behaviour, and mode collapse [49,50]. The bounded and gradually stabilising score trajectories observed in Fig. 6 indicate that neither network consistently dominated the optimisation process.

Time-domain comparisons are also performed to assess whether generated signals replicate the waveform characteristics of the real dataset. Representative real and generated signals are compared in terms of amplitude behaviour and temporal structure across both healthy and damaged conditions. The signal amplitude is also quantified using the root mean square, computed for each signal and averaged per class as:

$$\text{RMS}_i = \sqrt{\frac{1}{T}\sum_{n=1}^{T} x_i^2[n]} \tag{17}$$

where $\mu[n]$ denotes the ensemble mean signal amplitude at time step $n$:

$$\sigma[n] = \sqrt{\frac{1}{N-1}\sum_{i=1}^{N}(x_i[n] - \mu[n])^2} \tag{18}$$

$$\text{CV} = \frac{\frac{1}{T}\sum_{n=1}^{T}\sigma[n]}{\frac{1}{T}\sum_{n=1}^{T}|\mu[n]|} \tag{19}$$

Importantly, the variability is assessed by computing the standard deviation across signals at each time step and averaging over time to obtain a global measure of signal dispersion. To allow comparison across datasets, the coefficient of variation, CV, is used by normalising the standard deviation with the mean absolute signal amplitude. A power spectral density, PSD, analysis is then conducted to evaluate whether the generated signals can reproduce the frequency content of the measured response.

$$\text{RMSE}_{\text{PSD}} = \sqrt{\frac{1}{F}\sum_{f=f_1}^{f_2}\left(PSD_{\text{real}}(f) - PSD_{\text{gen}}(f)\right)^2} \tag{20}$$

The quantitative comparison is performed using the root mean square error RMSE between the mean PSDs of real and generated signals. RMSE is evaluated across the primary excitation bandwidth 5–256 Hz, as well as higher frequencies up to the Nyquist limit of 1024 Hz, enabling assessment of both dominant dynamics and higher-order spectral behaviour. To complement this, the spectral energy within a selected frequency range is computed by integrating the PSD:

$$E = \int_{f_1}^{f_2} PSD(f)\,df \tag{21}$$

where, $f_1$ and $f_2$ denote the lower and upper frequency limits of the considered frequency band. The ratio between generated and real spectral energy is then evaluated as:

$$R_E = \frac{E_{\text{gen}}}{E_{\text{real}}} \tag{22}$$

where, $E_{\text{gen}}$ and $E_{\text{real}}$ are the spectral energies of the generated and measured signals, respectively.

The principal component analysis, PCA, is also applied to project these features into a lower-dimensional space defined by the dominant modes of variation. Let $\mathbf{X} \in \mathbb{R}^{n\times d}$ denote the feature matrix, where $n$ is the number of samples and $d$ is the number of extracted features. PCA projects the feature matrix onto a lower-dimensional space according to:

$$\mathbf{Z} = \mathbf{X}\mathbf{W} \tag{23}$$

where, $\mathbf{X} \in \mathbb{R}^{n\times d}$ denotes the data matrix, $\mathbf{W}$ contains the principal component loading vectors, and $\mathbf{Z}$ is the matrix of projected features. The proportion of variance explained by the $k$th principal component is given by:

$$\eta_k = \frac{\lambda_k}{\sum_{j=1}^{d}\lambda_j} \tag{24}$$

where, $\lambda_k$ is the $k$th eigenvalue of the covariance matrix of $\mathbf{X}$ and $d$ is the total number of components. The first three principal components are examined to enable visualisation of the feature space while retaining most of the variance in the real dataset. Finally, a support vector machine classifier is applied to evaluate accuracy, precision, recall, and F1-score as:

$$\text{Accuracy} = \frac{TP+TN}{TP+TN+FP+FN} \tag{25}$$

$$\text{Precision} = \frac{TP}{TP+FP} \quad \text{Recall} = \frac{TP}{TP+FN} \tag{26}$$

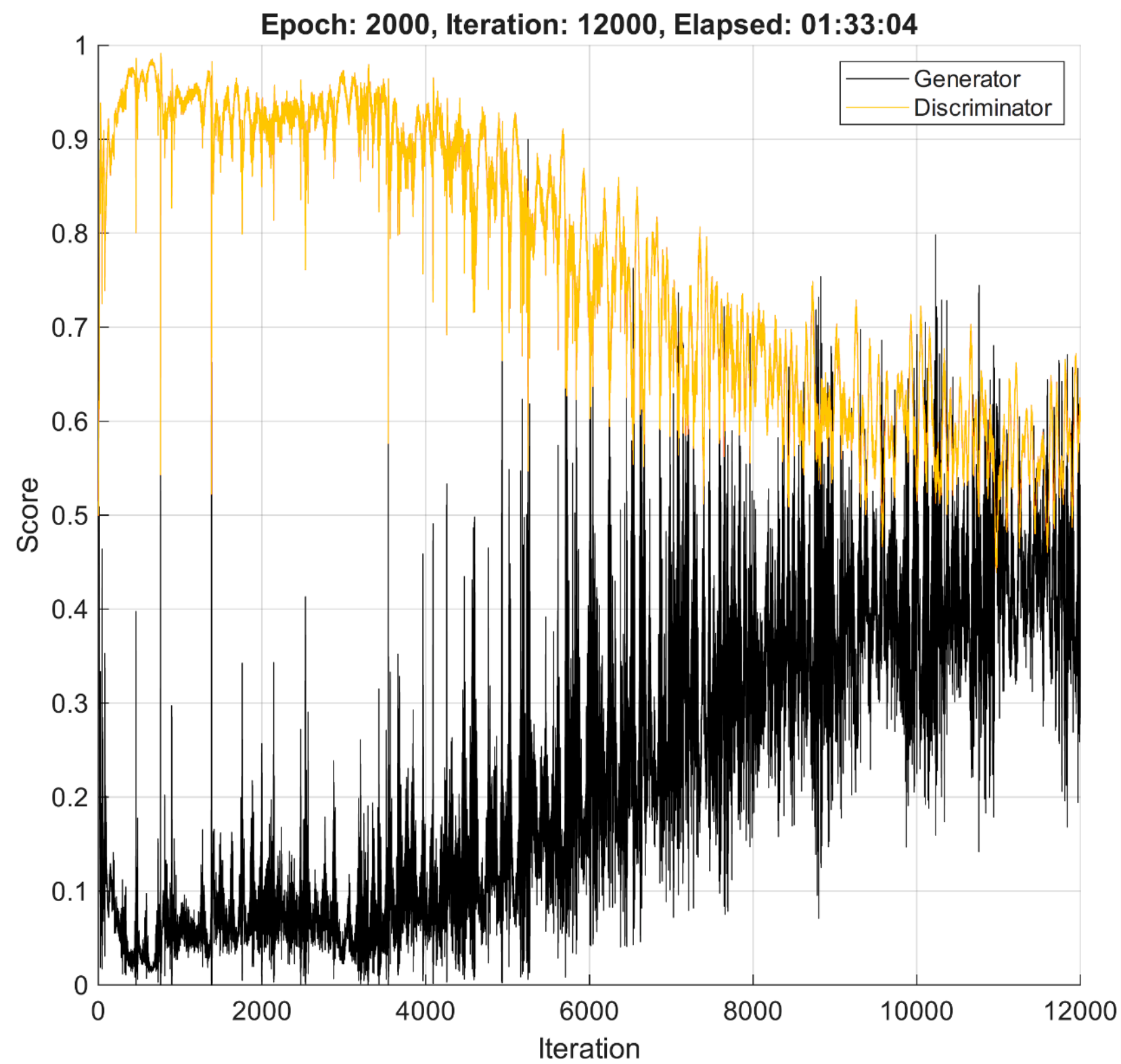

**Fig. 6.** Evolution of generator and discriminator scores during adversarial training for the WN03-A dataset.

$$F_1 = 2 \cdot \frac{\text{Precision} \cdot \text{Recall}}{\text{Precision} + \text{Recall}} \quad (27)$$

For feature-space visualisation, each vibration signal is represented using a 14-dimensional feature vector derived from both spectral and time-domain characteristics, seen in Table 8. The extracted features included the dominant spectral peak frequency, spectral energy in selected frequency bands, spectral kurtosis peak frequency, mean value, variance, skewness, kurtosis, peak-to-peak amplitude, crest factor, RMS amplitude, median absolute deviation, and cumulative-sum range.

To improve reproducibility, Table 9 summarises the implementation settings used for PCA, SVM classification, synthetic data generation, and hyperparameter selection.

**Table 8**
Features extracted from each vibration signal for PCA visualisation and SVM classification.

| Feature | Description |
|---|---|
| $f_{\text{peak}}$ | Frequency corresponding to the maximum PSD amplitude |
| $P_{\text{low}}$ | Spectral energy within the 10–20 Hz band |
| $P_{\text{mid}}$ | Spectral energy within the 40–60 Hz band |
| $P_{\text{high}}$ | Spectral energy above 100 Hz |
| $f_{\text{SK}}$ | Frequency corresponding to the maximum spectral kurtosis |
| $\mu$ | Signal mean |
| $\sigma^2$ | Signal variance |
| $\gamma$ | Signal skewness |
| $\kappa$ | Signal kurtosis |
| $A_{\text{pp}}$ | Peak-to-peak amplitude |
| $C_{\text{f}}$ | Crest factor (peak-to-RMS ratio) |
| RMS | Root mean square amplitude |
| MAD | Median absolute deviation |
| $R_{\text{CS}}$ | Range of the cumulative sum of the signal |

## 5. Application to fibre Bragg grating strain, acceleration and nonlinear dynamic response measurements

The performance of the approach is examined under varying excitation conditions, damage states, and sensor types. The time-domain responses (Fig. 7a,b) show that both real and generated signals exhibit zero-mean, broadband behaviour characteristic of multisine excitation. The generated signals reproduce the overall signal structure and amplitude envelope of the measured response, supported by the close agreement in RMS amplitude (Table 11).

The power spectral density shows strong agreement between real and generated signals within the primary excitation band of 5–256 Hz (Fig. 7c,d). In this region, low error values and energy ratios slightly above unity (Table 10) indicate that the model accurately captures the dominant frequency characteristics of the structural response, with a mild overestimation of spectral energy. Although excitation is limited to 5–256 Hz, measurable spectral content beyond this range reflects residual structural response at higher frequencies due to nonlinear effects and measurement noise. In the intermediate range of 256–512 Hz, agreement remains good, with only a small increase in the error and energy ratios close to unity (Table 10), indicating that the model captures secondary spectral features with slightly reduced accuracy. Here, the primary excitation bandwidth 5–256 Hz is evaluated separately as it contains the dominant structural response. Additional subdivision of the higher-frequency region is introduced only in cases where noticeable spectral deviations are observed beyond the excitation range. For the strain datasets, the 256–1024 Hz range is separated into 256–512 Hz and 512–1024 Hz bands to distinguish intermediate-frequency behaviour from the larger discrepancies observed at the highest frequencies. In contrast, the acceleration datasets exhibited a more uniform high-frequency response and are therefore assessed using a single 256–1024 Hz band.

In the high-frequency range 512–1024 Hz, a clear divergence emerges. The real signals exhibit a sharp spectral decay, whereas the generated signals retain elevated spectrum levels and decay more gradually (Fig. 7c,d). As shown in Table 10, this results in larger errors, with error values exceeding 90 dB and energy ratios indicating signifi-cant overestimation of spectral energy, while the low-mid frequencies' exact pointwise agreement between stochastic signal realisations is not expected or intended.

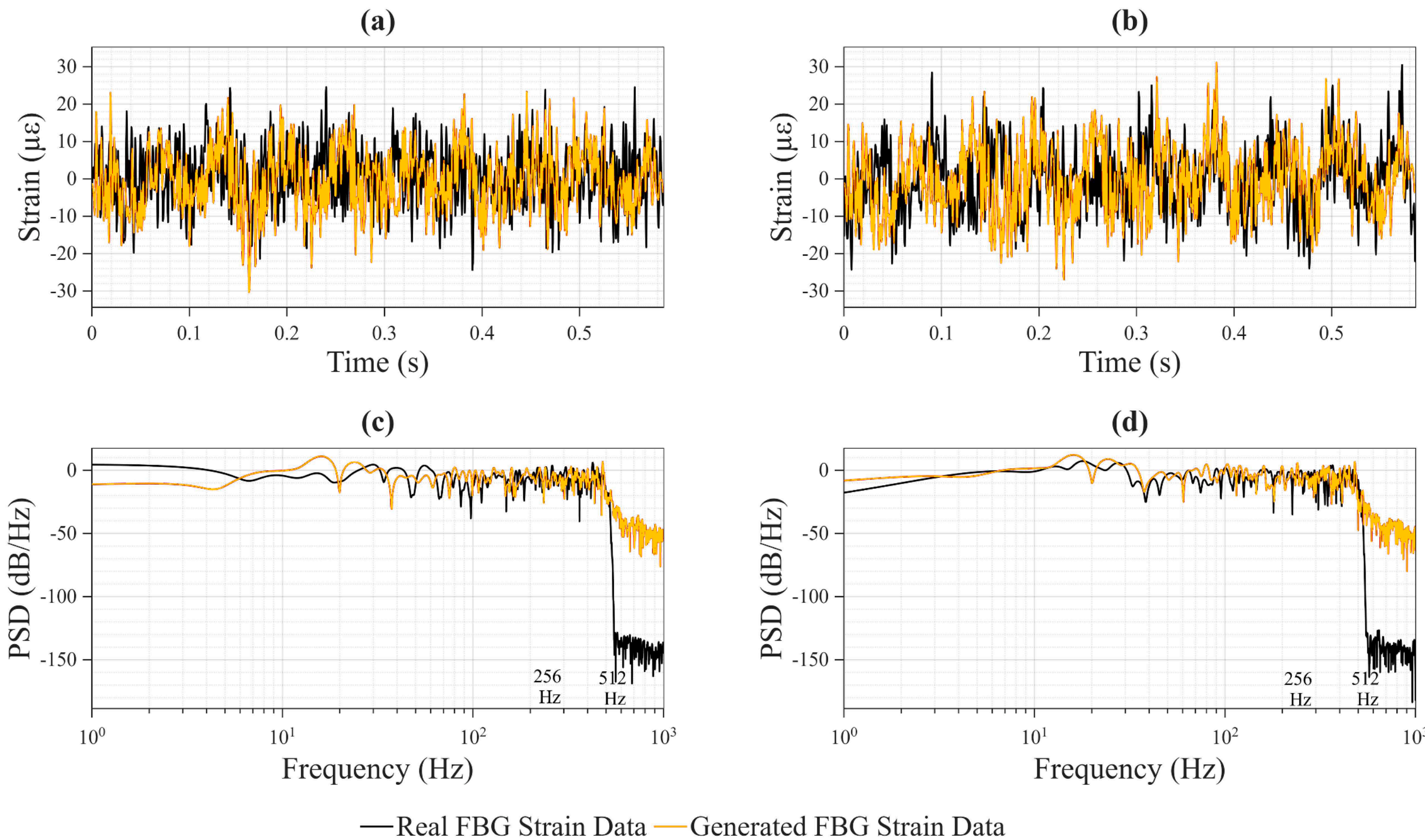

**Fig. 7.** Comparison of representative real and cGAN generated signals for the ORPM02-S dataset: (a) healthy state, time domain (b) damaged state, time domain (c) healthy state, spectral domain and (d) damaged state, spectral domain.

**Table 9**
Implementation details for PCA, SVM classification, and data augmentation.

| Item | Setting |
|---|---|
| Feature representation | 14 statistical and spectral features (Table 8). |
| PCA method | Principal component analysis using the extracted feature matrix. |
| Classifier | Support Vector Machine (SVM). |
| Kernel function | Second-order polynomial kernel. |
| Polynomial order | 2. |
| Kernel scale | Automatic. |
| Box constraint | 1.0. |
| Input standardisation | Enabled prior to training. |
| Train–test split | 70/30 hold-out split. |
| Generated samples | 2000 synthetic samples (1000 healthy and 1000 damaged). |
| Augmentation ratio | 1:1 ratio between generated healthy and generated damaged samples. |
| Hyperparameter selection | Selected empirically through preliminary experiments to achieve stable adversarial training, convergence of generator and discriminator scores, and good agreement between real and generated signals. |

**Table 10**
Frequency-domain comparison of real and cGAN generated signals for healthy and damaged states in the ORPM02-S dataset.

| Metric | Frequency Range | Healthy | Damaged |
|---|---|---|---|
| RMSE (dB) | 5–256 Hz | 4.67 | 4.56 |
| | 256–512 Hz | 5.54 | 5.27 |
| | 512–1024 Hz | 90.98 | 91.29 |
| Energy ratio (Gen/Real) | 5–256 Hz | 1.16 | 1.13 |
| | 256–512 Hz | 0.99 | 1.07 |
| | 512–1024 Hz | 2.63 | 2.97 |

**Table 11**
Comparison of signal amplitude (RMS) and inter-signal variability (standard deviation and coefficient of variation) for real and cGAN generated signals in the ORPM02-S dataset.

| Metric | Condition | Real | Generated | % Difference |
|---|---|---|---|---|
| RMS amplitude | Healthy | 8.54 | 8.75 | 2.46 |
| RMS amplitude | Damaged | 8.83 | 9.11 | 3.17 |
| Standard deviation | – | 8.69 | 3.06 | −64.79 |
| Coefficient of variation | – | 1.25 | 0.60 | −52.00 |

A reduction in variability is also observed in the generated signals (Table 11), with the inter-signal standard deviation decreasing by approximately 65% relative to the real data. A similar reduction is observed in the coefficient of variation, although to a lesser extent due to the higher mean amplitude of the generated signals.

A strong agreement between the distributions of real and generated signals is observed in the reduced feature space (Fig. 8), with the first three principal components capturing 90.25% of the total variance. The generated signals show substantial overlap with the real data, indicating that the model reproduces the overall statistical structure of the vibration response. However, the generated signals form a more compact cluster than the real data, reflecting reduced variance. This is consistent with the previously observed reduction in inter-signal variability (Table 11). The inclusion of model-generated data results in a small improvement in the support vector machine performance across all metrics (Table 12).

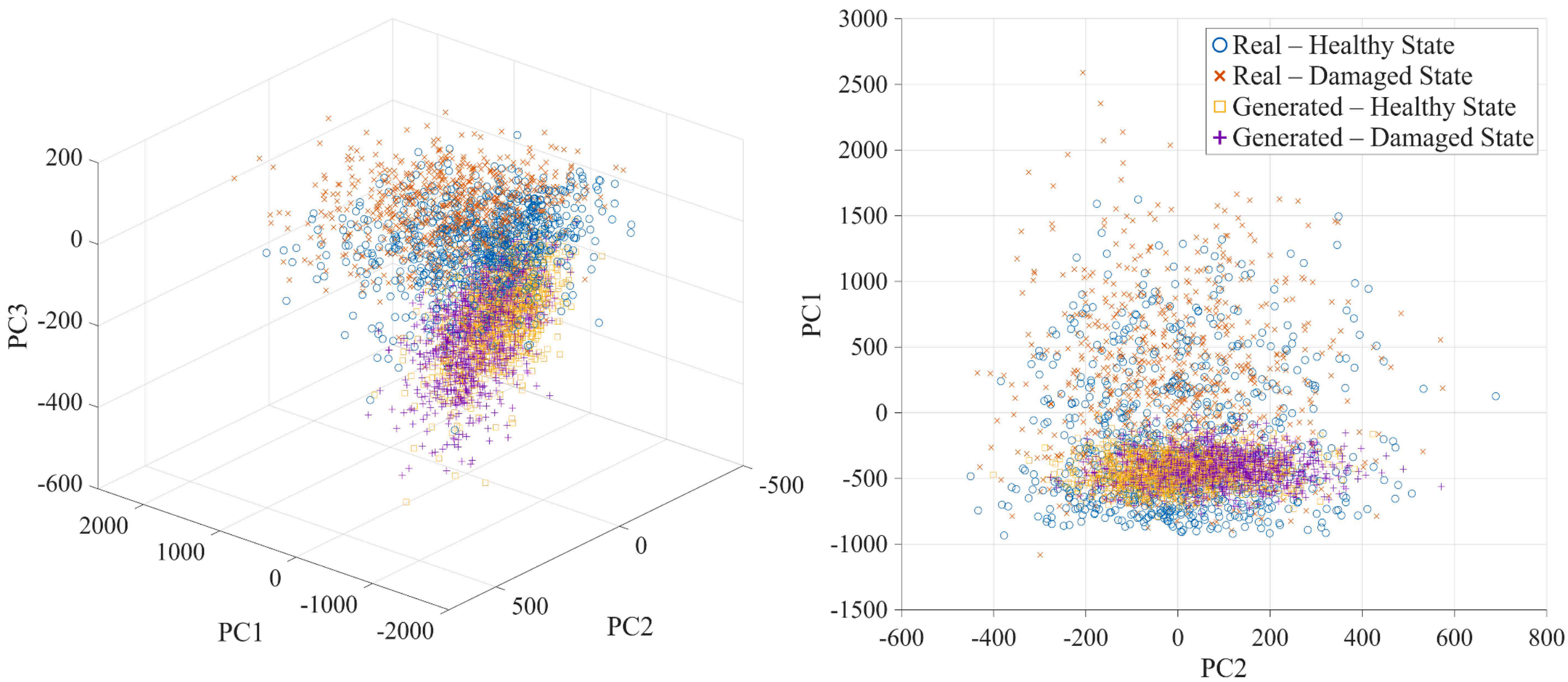

**Fig. 8.** PCA feature space representation of extracted vibration features for real and cGAN generated signals for the ORPM02-S dataset, projected onto the first three principal components for healthy and damaged states.

**Table 12**
SVM classification performance on the ORPM02-S dataset for models trained on real and cGAN generated data, evaluated on real signals.

| Training Dataset | Accuracy (%) | Precision (%) | Recall (%) | F1 Score (%) |
|---|---|---|---|---|
| Real (Baseline) | 80.78 ± 1.74 | 82.06 ± 2.91 | 78.63 ± 2.87 | 80.24 ± 1.80 |
| Real Generated | 81.13 ± 1.77 | 82.14 ± 2.67 | 79.35 ± 2.62 | 80.68 ± 1.77 |

**Table 13**
Frequency-domain comparison of real and cGAN generated signals for the ORPM02-A dataset.

| Metric | Frequency Range | Healthy | Damaged |
|---|---|---|---|
| RMSE (dB) | 5–256 Hz | 6.06 | 6.14 |
| | 256–1024 Hz | 7.45 | 7.10 |
| Energy ratio (Gen/Real) | 5–256 Hz | 1.10 | 1.06 |
| | 256–1024 Hz | 1.08 | 1.37 |

The uncertainty in the generated signals is quantified using Monte Carlo dropout. A normalised uncertainty of 0.446 is observed for both healthy and damaged states.

The ORPM02-A dataset is also used to assess the model's performance under the same experimental configuration, with acceleration measured on the port wing. In the time domain, both real and generated signals exhibit zero-mean, broadband behaviour consistent with multisine excitation (Fig. 9a,b). The generated signals capture the overall temporal behaviour and amplitude characteristics of the measured response, supported by the close agreement in RMS amplitude (Table 14).

The frequency-domain analysis shows strong agreement between real and generated signals across the dominant frequency range (Fig. 9c,d). The overall spectrum shape and energy distribution are well preserved within the primary excitation band 5–256 Hz, with low error values and energy ratios close to unity, indicating accurate reproduction of the dominant spectral response (Table 13). In the higher frequency range 256–1024 Hz, a modest increase in error is observed, with energy ratios slightly above unity in the healthy state and a greater overestimation in the damaged state.

A reduction in inter-signal variability is also observed in the generated signals relative to the real data (Table 14), as reflected by the lower standard deviation and coefficient of variation. This indicates a narrower spread in the generated responses.

**Table 14**
Comparison of signal amplitude (RMS) and inter-signal variability (standard deviation and coefficient of variation) for real and cGAN generated signals in the ORPM02-A dataset.

| Metric | Condition | Real | Generated | % Difference |
|---|---|---|---|---|
| RMS amplitude | Healthy | 0.321 | 0.327 | 1.87 |
| RMS amplitude | Damaged | 0.514 | 0.516 | 0.39 |
| Standard deviation | – | 0.429 | 0.359 | −16.32 |
| Coefficient of variation | – | 1.27 | 1.09 | −14.17 |

In the principal component analysis space, the generated signals closely follow the distribution of the real data (Fig. 10), with the first three principal components capturing 98.15% of the total variance. This indicates that the dominant signal characteristics are well represented within the reduced feature space. The generated samples overlap with the real data, indicating that the model captures the primary statistical structure of the acceleration response. The feature space exhibits clustering, with real and generated signals forming similarly structured patterns. However, the generated signals occupy slightly more concentrated regions, indicating reduced spread compared to the real data. This is consistent with the reduced variability observed in Table 14. A small number of generated samples lie outside the main cluster (Fig. 10), indicating occasional deviations from the learned data distribution, with some samples not fully aligned with the real data feature space. A separation between healthy and damaged states is observed, with real and generated signals forming distinct clusters associated with each condition (Fig. 10). Within each state, the generated samples occupy the same regions as the real data.

A high support vector machine performance is observed across both training configurations (Table 15) reflecting the high degree of class separability in this dataset. The slight reduction in accuracy suggests that the generated data slightly reduces the separation between classes, with the highly distinct boundaries observed in the real data (Fig. 10) becoming less pronounced.

The normalised uncertainty is similar for both healthy and damaged states, with values of 0.327 and 0.342, respectively. This near-uniformity indicates a consistent level of stochastic variation in the generated signals, with no dependence on structural state. This behaviour is also observed in Fig. 11, where the uncertainty bounds exhibit a similar spread for both conditions. Fig. 11 is presented as a representative example of the Monte Carlo dropout uncertainty analysis.

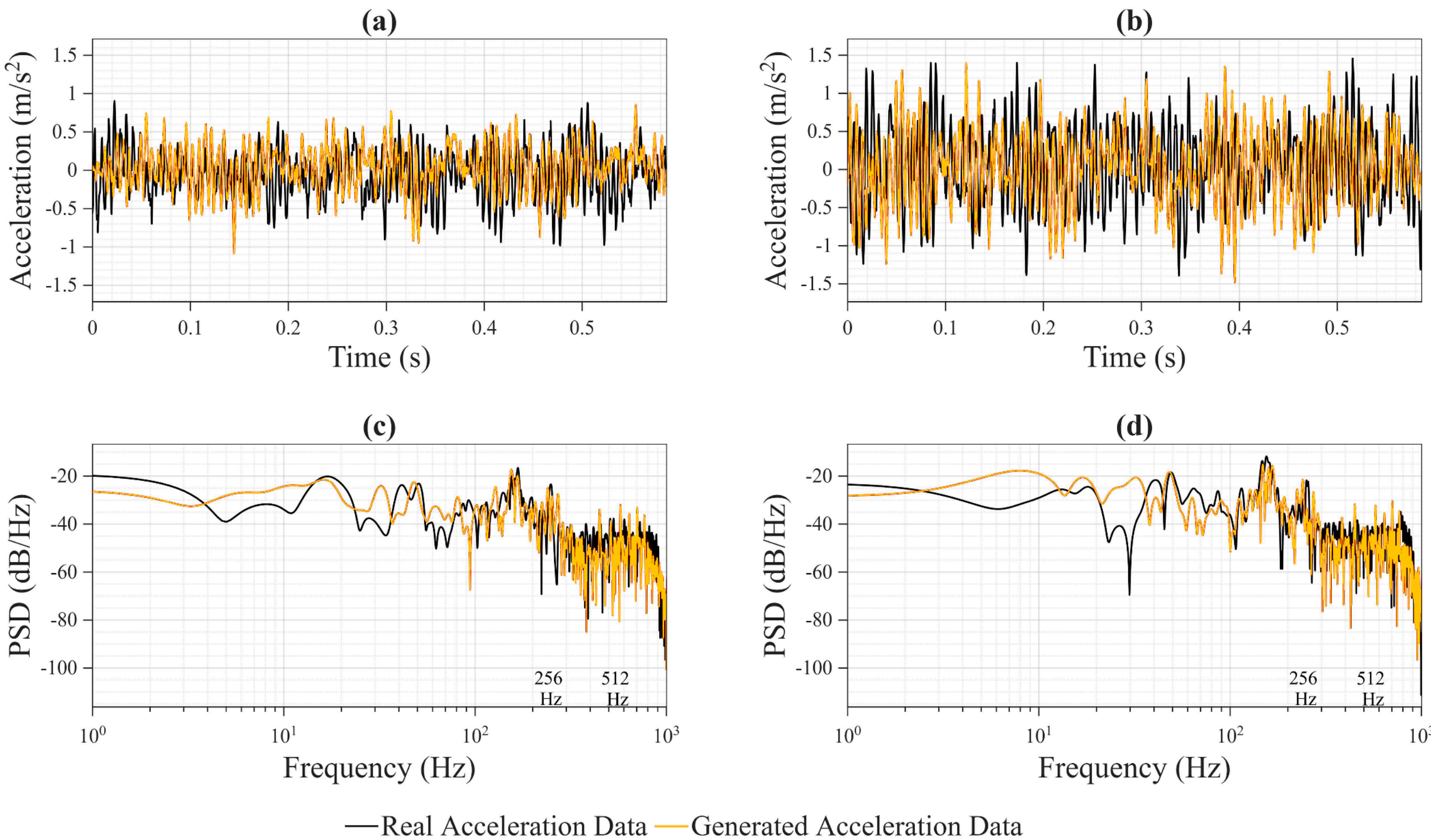

**Fig. 9.** Comparison of representative real and cGAN generated signals for the ORPM02-A dataset: (a) healthy state, time domain (b) damaged state, time domain (c) healthy state, spectral domain and (d) damaged state, spectral domain.

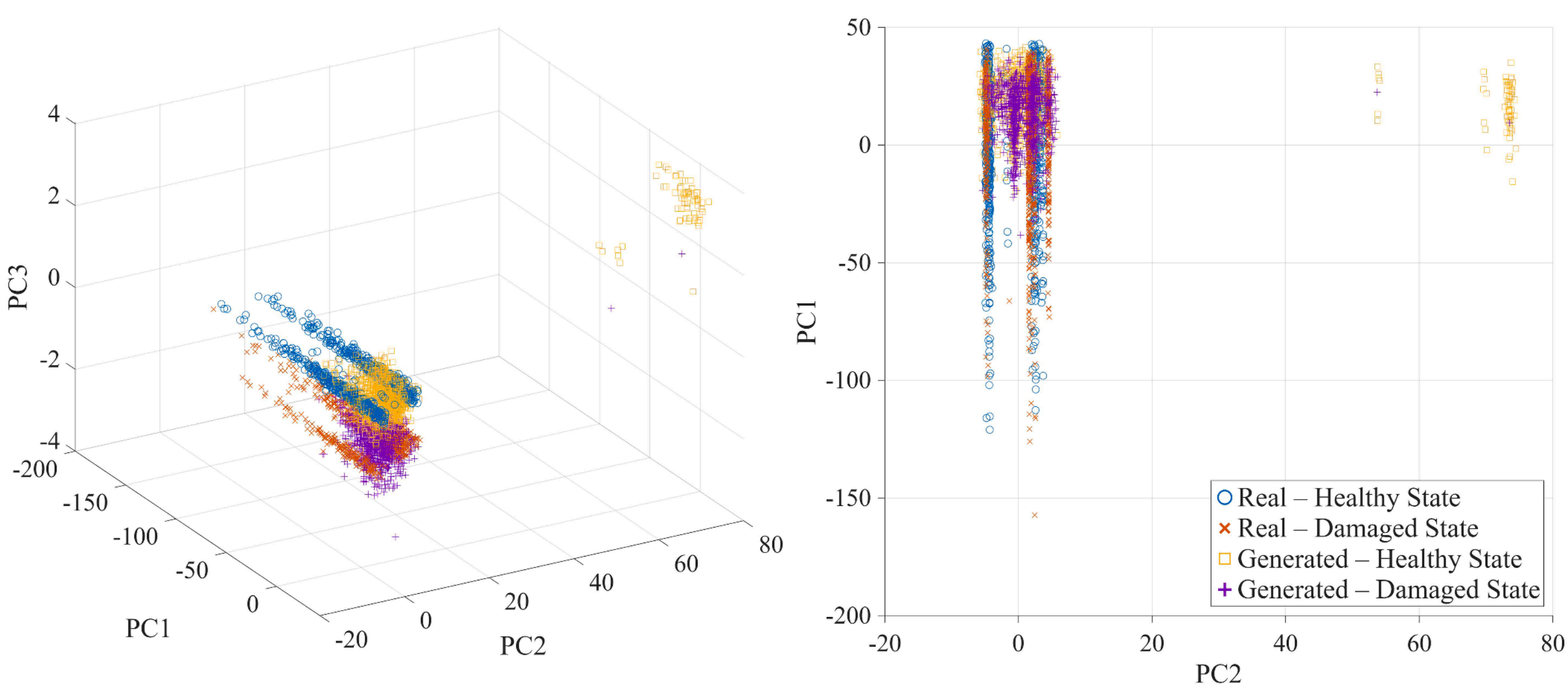

**Fig. 10.** PCA feature-space representation of extracted vibration features for real and cGAN generated signals for the ORPM02-A dataset, projected onto the first three principal components for healthy and damaged states.

The model is also evaluated on the WN03-S dataset. In the time domain, both real and generated signals exhibit zero-mean, broadband behaviour characteristic of white noise excitation (Fig. 12a,b). The generated signals capture the irregular and rapidly varying nature of the measured response, reproducing key features such as local amplitude variations. This is supported by the close agreement in root mean square amplitude (Table 17), indicating accurate reproduction of overall signal magnitude for both structural states.

In the frequency domain, the power spectral density shows that the generated signals retain the main spectral frequency bands of the real data within the primary excitation band of 5–256 Hz (Fig. 12c,d). Within this range, the overall energy distribution and dominant frequency components are well preserved, with low error values and energy ratios slightly above unity (Table 16), indicating a moderate overestimation of spectral energy. In the intermediate range of 256–512 Hz, agreement remains good, with low error and energy ratios close to unity, although a slight underestimation of spectral energy is observed. This indicates that the model captures both primary and secondary spectral features with similar accuracy to the ORPM02-S dataset.

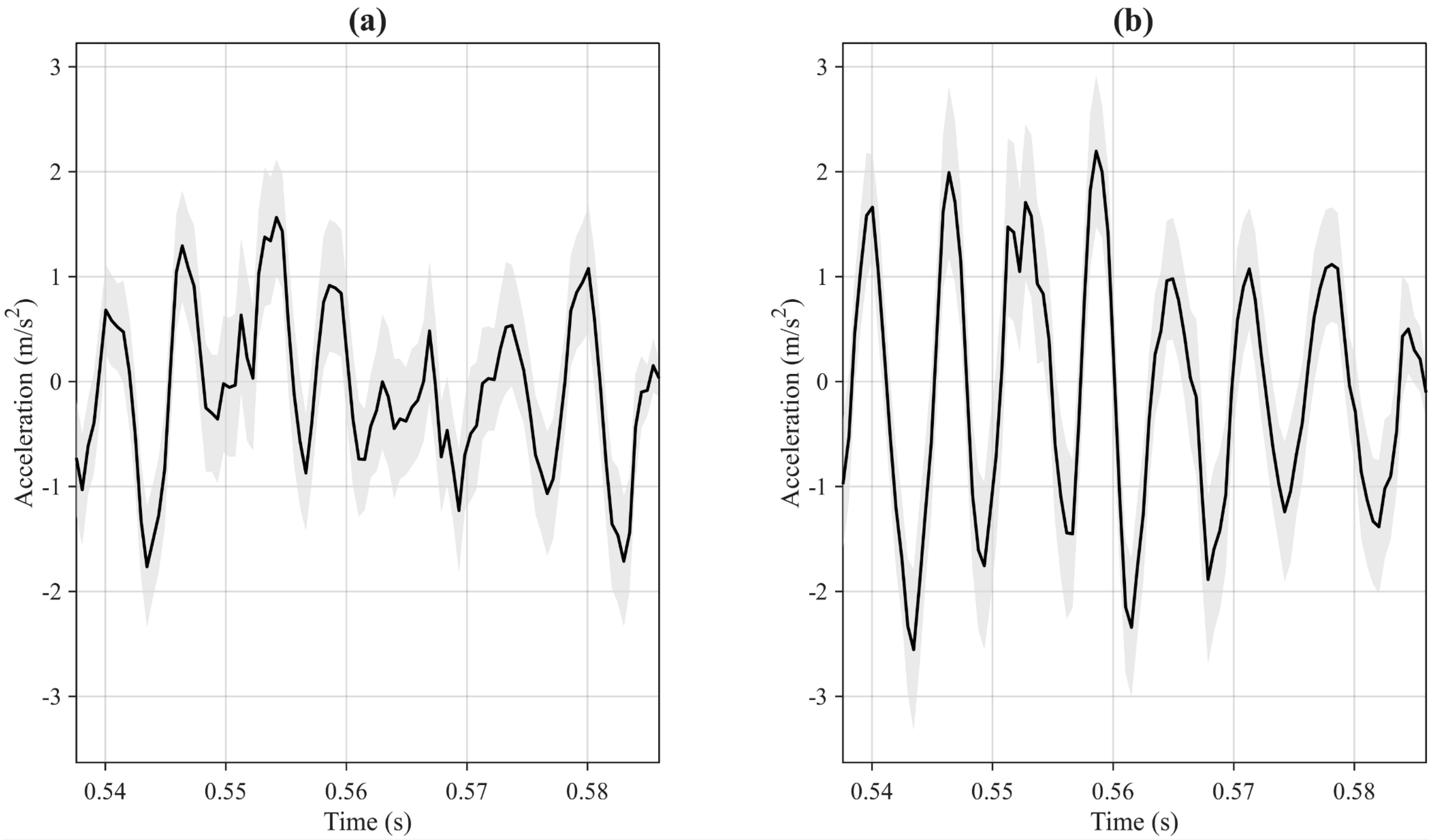

**Fig. 11.** Monte Carlo generator uncertainty for the ORPM02-A dataset.

**Table 15**
SVM classification performance on the ORPM02-A dataset for models trained on real and cGAN generated data, evaluated on real signals.

| Training Dataset | Accuracy (%) | Precision (%) | Recall (%) | F1 Score (%) |
|---|---|---|---|---|
| Real (Baseline) | 99.94 ± 0.02 | 100.00 ± 0.00 | 99.87 ± 0.04 | 99.94 ± 0.02 |
| Real Generated | 99.92 ± 0.06 | 99.99 ± 0.09 | 99.86 ± 0.09 | 99.92 ± 0.06 |

**Table 16**
Frequency-domain comparison of real and cGAN generated signals for the WN03-S dataset.

| Metric | Frequency Range | Healthy | Damaged |
|---|---|---|---|
| RMSE (dB) | 5–256 Hz | 4.37 | 4.25 |
| | 256–512 Hz | 4.61 | 4.68 |
| | 512–1024 Hz | 87.30 | 87.44 |
| Energy ratio (Gen/Real) | 5–256 Hz | 1.29 | 1.16 |
| | 256–512 Hz | 0.90 | 0.96 |
| | 512–1024 Hz | 1.76 | 1.96 |

**Table 17**
Comparison of signal amplitude (RMS) and inter-signal variability (standard deviation and coefficient of variation) for real and cGAN generated signals in the WN03-S dataset.

| Metric | Condition | Real | Generated | % Difference |
|---|---|---|---|---|
| RMS amplitude | Healthy | 7.84 | 8.13 | 3.70 |
| RMS amplitude | Damaged | 8.22 | 8.38 | 1.93 |
| Standard deviation | – | 7.96 | 3.21 | −59.60 |
| Coefficient of variation | – | 1.24 | 0.66 | −47.16 |

In the high-frequency range of 512–1024 Hz, a divergence between real and generated signals is observed. The real signals exhibit a rapid decay in spectral energy, whereas the generated signals retain elevated energy levels. This is reflected in Table 16, where error values increase substantially and energy ratios exceed unity, indicating overestimation of high-frequency content. However, this discrepancy is confined to a low-energy region of the spectrum, with the dominant structural response remaining well represented. A similar behaviour is observed for the ORPM02-S dataset, indicating that high-frequency deviations are consistent across excitation types while the primary response is accurately captured.

A significant reduction in variability is observed in the generated signals (Table 17), with the inter-signal standard deviation decreasing by approximately 60% and the coefficient of variation by 47% relative to the real data.

The projection onto the first three principal components retains 89.65% of the total variance (Fig. 13), indicating that the dominant signal characteristics are well represented within the reduced feature space. The generated signals broadly follow the distribution of the real data, suggesting that the overall statistical structure is captured. However, the feature space is characterised by substantial overlap between healthy and damaged states, indicating limited class separability within the dataset. The generated samples follow this behaviour closely, occupying similar regions while forming slightly more concentrated clusters, consistent with the reduced variability observed in Table 17.

The low support vector machine performance observed across both training configurations (Table 18) reflects the limited class separability in this dataset, indicating a more challenging classification scenario. In contrast to the previous cases, the inclusion of the model-generated data results in a slight reduction in performance, particularly in recall and F1 score, suggesting that the generated samples do not fully capture the class-discriminative features.

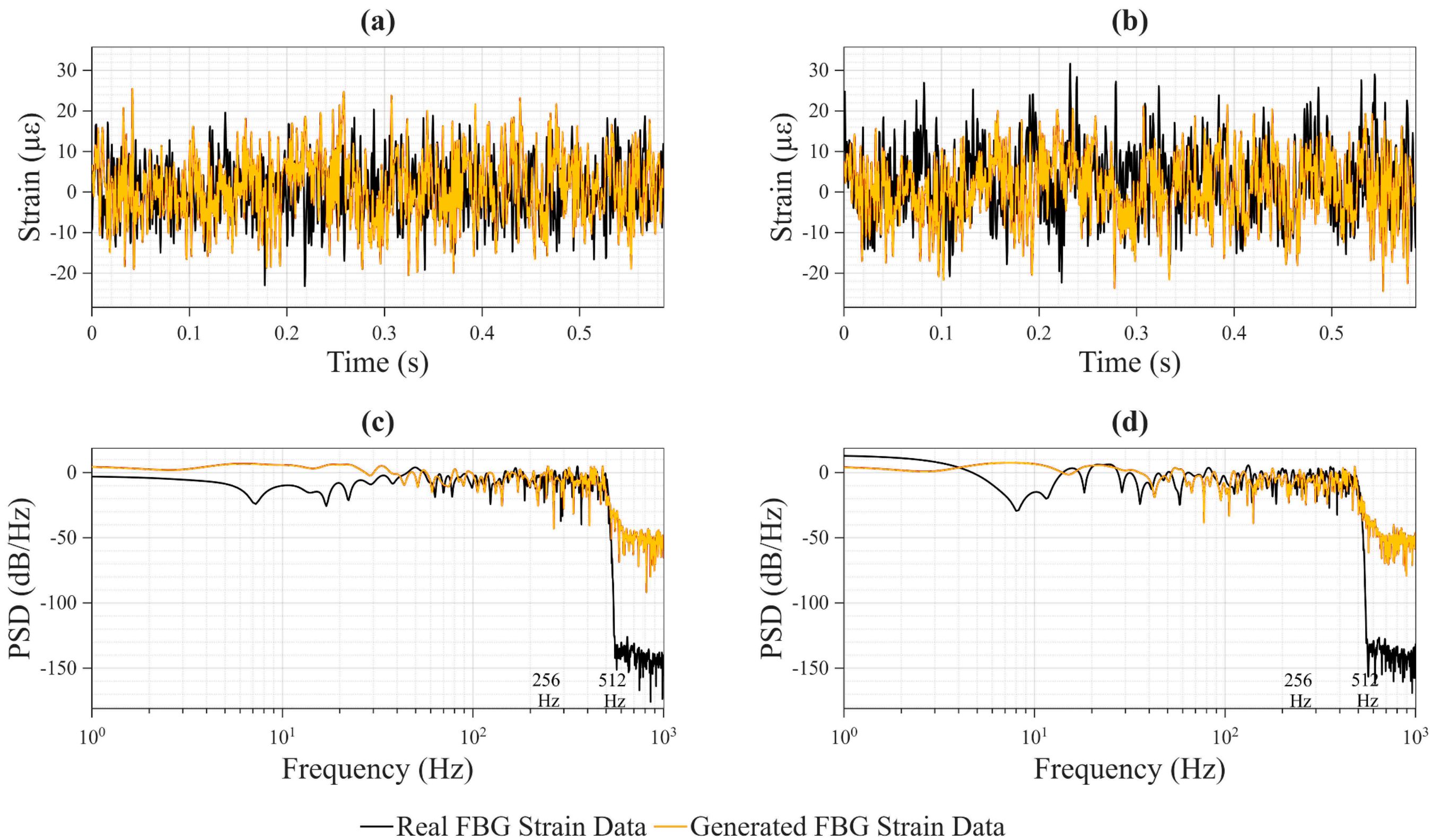

**Fig. 12.** Comparison of representative real and cGAN generated signals for the WN03-S dataset: (a) healthy state, time domain (b) damaged state, time domain (c) healthy state, spectral domain and (d) damaged state, spectral domain.

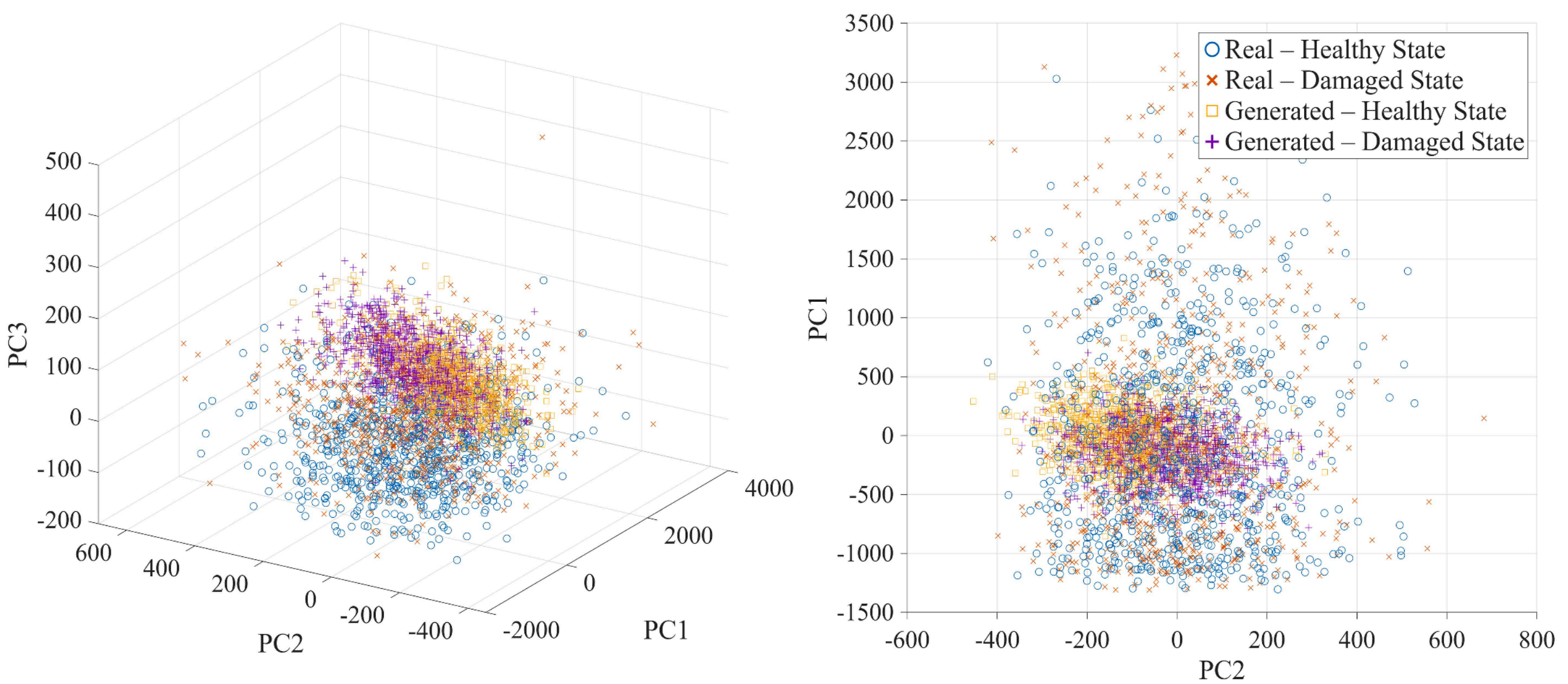

**Fig. 13.** PCA feature-space representation of extracted vibration features for real and cGAN generated signals for the WN03-S dataset, projected onto the first three principal components for healthy and damaged states.

The normalised uncertainty is 0.521 and 0.524 for the healthy and damaged states, respectively, with only a small difference between conditions. The higher uncertainty, relative to the ORPM02-S dataset, indicates that the generated signals exhibit less consistent structure, suggesting that the WN03-S dataset is more challenging for the model to represent accurately, consistent with the more stochastic nature of white noise excitation. The results show that the model captures the dominant spectral and statistical characteristics of the strain response under white noise excitation. However, reduced variability, overestimation of high-frequency content, and substantial class overlap indicate limitations in fully representing the underlying data distribution. These limitations are reflected in the reduced classification performance, highlighting the difficulty of modelling complex and less separable structural responses.

A final evaluation of the model performance is conducted using the WN03-A dataset. The acceleration signals exhibit the expected zero-mean, broadband stochastic behaviour under white noise excitation, with the generated outputs closely matching the measured response (Fig. 14a,b). Key features such as amplitude variations are well captured, indicating close agreement with the measured response. This is supported by the close agreement in root mean square amplitude across both structural states (Table 20), indicating accurate reproduction of overall signal magnitude.

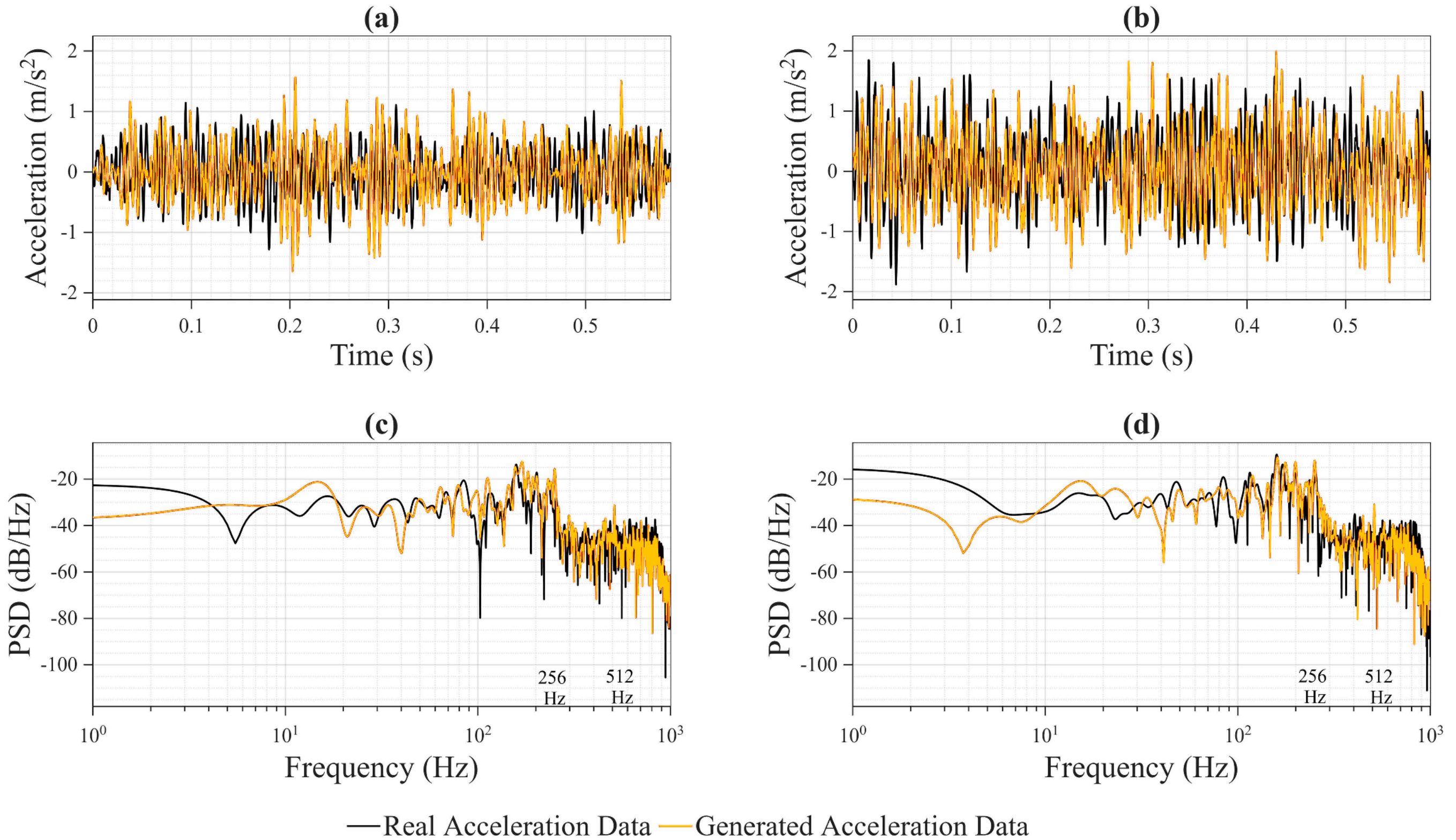

**Fig. 14.** Comparison of representative real and cGAN generated signals for the WN03-A dataset: (a) healthy state, time domain (b) damaged state, time domain (c) healthy state, spectral domain and (d) damaged state, spectral domain.

**Table 18**
SVM classification performance on the WN03-S dataset for models trained on real and cGAN generated data, evaluated on real signals.

| Training Dataset | Accuracy (%) | Precision (%) | Recall (%) | F1 Score (%) |
|---|---|---|---|---|
| Real (Baseline) | 67.27 ± 2.07 | 68.40 ± 3.36 | 65.07 ± 5.03 | 66.51 ± 2.43 |
| Real Generated | 66.13 ± 2.02 | 68.10 ± 3.44 | 61.53 ± 4.62 | 64.47 ± 2.48 |

**Table 19**
Frequency-domain comparison of real and cGAN generated signals for the WN03-A dataset.

| Metric | Frequency Range | Healthy | Damaged |
|---|---|---|---|
| RMSE (dB) | 5–256 Hz | 3.57 | 3.57 |
| | 256–1024 Hz | 7.40 | 6.78 |
| Energy ratio (Gen/Real) | 5–256 Hz | 1.15 | 1.03 |
| | 256–1024 Hz | 0.57 | 0.84 |

**Table 20**
Comparison of signal amplitude (RMS) and inter-signal variability (standard deviation and coefficient of variation) for real and cGAN generated signals in the WN03-A dataset.

| Metric | Condition | Real | Generated | % Difference |
|---|---|---|---|---|
| RMS amplitude | Healthy | 0.413 | 0.418 | 1.02 |
| RMS amplitude | Damaged | 0.676 | 0.665 | −1.63 |
| Standard deviation | – | 0.561 | 0.450 | −19.82 |
| Coefficient of variation | – | 1.29 | 1.09 | −15.75 |

The generated signals retain the dominant low-frequency spectral profile and principal energy distribution, although deviations are observed at higher frequencies (Fig. 14c,d). The distribution of energy is well represented, indicating that the model captures the dominant response characteristics across the spectrum. This is supported by Table 19, where low error values are observed across both frequency ranges, with energy ratios close to unity in the primary band 5–256 Hz, showing a slight overestimation of spectral energy. At higher frequencies, however, energy ratios fall below unity, indicating an underestimation of spectral energy. This contrasts with the ORPM02-A dataset, where high-frequency energy is overestimated, highlighting a difference in how the model represents higher-frequency content under different excitation conditions.

A reduction in inter-signal variability is observed in the generated signals (Table 20), with the mean standard deviation decreasing by approximately 20% and the coefficient of variation by 16% relative to the real data. This indicates a slight reduction in variability between signals, suggesting that while the overall signal behaviour is preserved, finer stochastic features of the structural response are not fully captured.

This reduction is notably less pronounced than in the WN03-S dataset, consistent with the behaviour observed in the ORPM datasets, where variability is also better preserved in acceleration than in strain measurements.

The projection onto the first three principal components retains 95.60% of the total variance (Fig. 15), indicating that the dominant variability is well represented in the reduced feature space. The generated signals largely overlap with the real data, showing that the overall statistical distribution is accurately captured.

The generated samples form slightly more compact clusters, indicating reduced variability, consistent with the narrower spread observed in Table 20. Despite this, clear separation between healthy and damaged states is maintained for both real and generated data, indicating strong class separability. The generated samples follow the same distributional structure as the real data, preserving the key class-discriminative features. This combination of reduced variability and strong class separation is consistent with the ORPM02-A dataset, indicating that the model preserves feature space structure in acceleration data while slightly compressing the distribution.

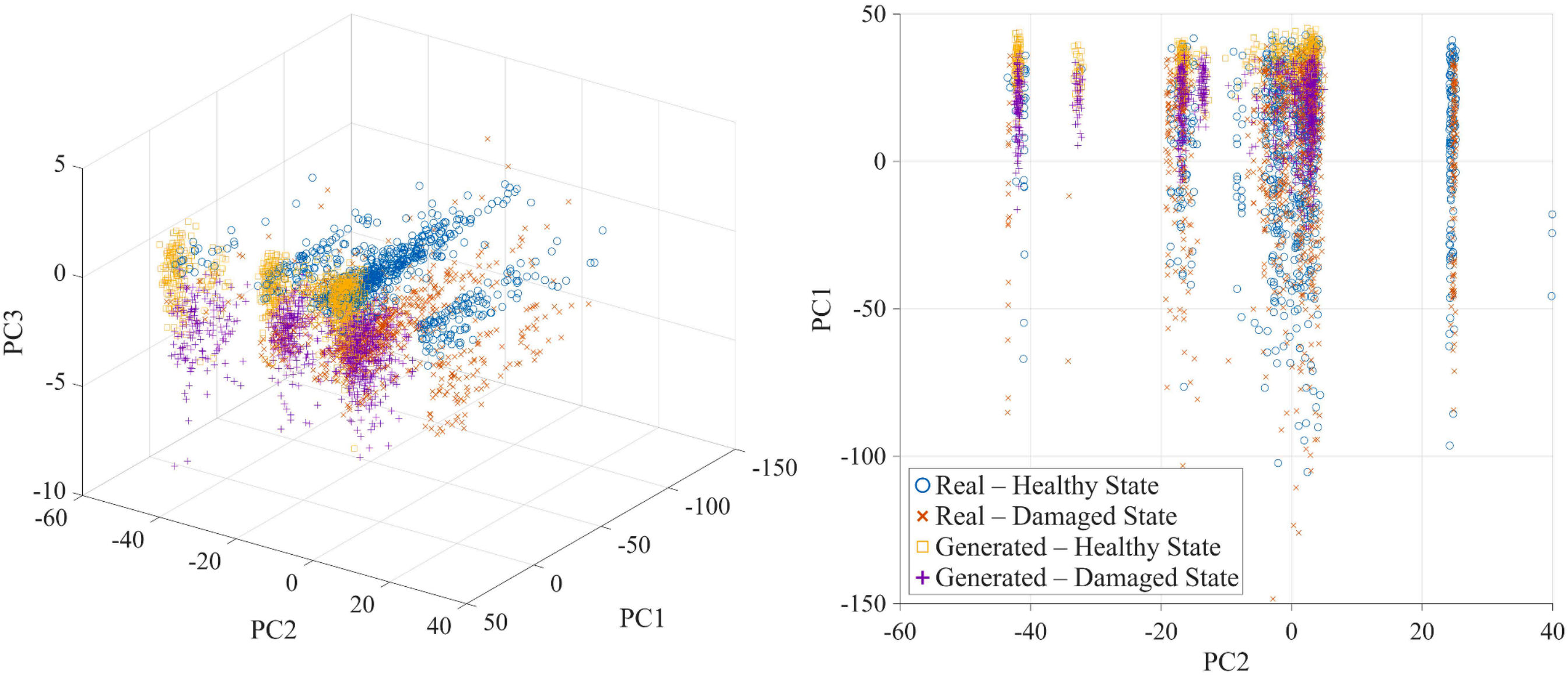

**Fig. 15.** PCA feature-space representation of extracted vibration features for real and cGAN generated signals for the WN03-A dataset, projected onto the first three principal components for healthy and damaged states.

**Table 21**
SVM classification performance on the WN03-A dataset for models trained on real and cGAN generated data, evaluated on real signals.

| Training Dataset | Accuracy (%) | Precision (%) | Recall (%) | F1 Score (%) |
|---|---|---|---|---|
| Real (Baseline) | 99.93 ± 0.04 | 99.98 ± 0.09 | 99.88 ± 0.05 | 99.93 ± 0.04 |
| Real Generated | 99.69 ± 0.29 | 99.81 ± 0.34 | 99.56 ± 0.46 | 99.69 ± 0.29 |

Similar to the high-separability previous case, the great support vector machine performance is observed across both training configurations (Table 21). However, the inclusion of model-generated data results in a slightly larger reduction in performance and increased variability compared to the ORPM case, although the magnitude of this change remains very small. This suggests that, while the model captures the overall class structure, it introduces minor inconsistencies in this dataset. Despite this, performance remains exceptionally high, indicating that strong class separability is still well preserved.

The normalised uncertainty is 0.434 and 0.464 for the healthy and damaged states, respectively, with only a small difference between conditions. While the variability remains consistent across structural states, these values are slightly higher than those reported for the ORPM02-A dataset. A similar trend is observed in the strain datasets, where the white noise case also exhibits higher uncertainty than its ORPM counterpart, indicating a consistent increase in uncertainty under white noise excitation.

## 6. Discussion

The examined framework was capable of reproducing characteristics of structural vibration signals with sparse sensing. However, this capability is not uniform across all conditions. While the model effectively represents the dominant, energy-driven behaviour of the system, it exhibits reduced ability in capturing the full variability of the data. This variation is reflected in the classification results. To assess signal usefulness, Fig. 16 is provided demonstrating that the uncertainty quantification provides further information. The datasets associated with lower uncertainty generally exhibit greater retention of the statistical variability present in the experimental measurements, whereas higher uncertainty is associated with stronger compression of the underlying data distribution. This behaviour suggests that the examined framework can act as a practical indicator of confidence in the generated responses, while also assessing whether generated samples are sufficiently representative for use in structural health monitoring tasks.

Along these lines, in the frequency domain, the model reproduces dominant response characteristics associated with the measured structural dynamics, which govern the overall system behaviour and are commonly used for vibration-based damage detection through changes in modal parameters such as natural frequencies and mode shapes [17]. However, this behaviour is not uniform across all frequency ranges. This behaviour reflects a bias towards learning the dominant components of the signal, which define its primary structure and have the greatest influence on reducing the training loss [51], leading to reduced sensitivity to finer-scale features. In contrast, the acceleration datasets show improved agreement at higher frequencies, with only modest deviations. This highlights that generative performance is strongly influenced by the nature of the measured response, with acceleration signals reflecting smoother, global structural dynamics, whereas strain signals capture more localised, higher-order behaviour and exhibit greater sensitivity to small-scale variations [52]. The model, exhibiting stable behaviour without evidence of mode collapse [53], is therefore well suited to applications requiring accurate representation of dominant structural dynamics, where primary frequency content and vibration-based features are central to structural health monitoring and damage detection [8]. However, early-stage damage and local defects often manifest as subtle variations in the structural response, as their small scale makes them difficult to detect using global vibration characteristics [54], which may not be accurately captured.

To further investigate the uncertainty and the variability on performance, Fig. 17 provides additional insight into the relationship between generated-data fidelity and classification performance. The left plot confirms that the strain datasets experience substantially larger reductions in variability than the acceleration datasets, indicating that local strain responses are more difficult for the generator to reproduce accurately. The middle plot shows that the effect of data augmentation on classifica-tion performance is generally modest and strongly dataset dependent. Most importantly, the right plot demonstrates that substantial reductions in variability do not necessarily translate into equivalent reductions in classification performance. This suggests that preservation of class-discriminative information is more important than preservation of the complete statistical dispersion of the measured response which is also summarised in Table 22.

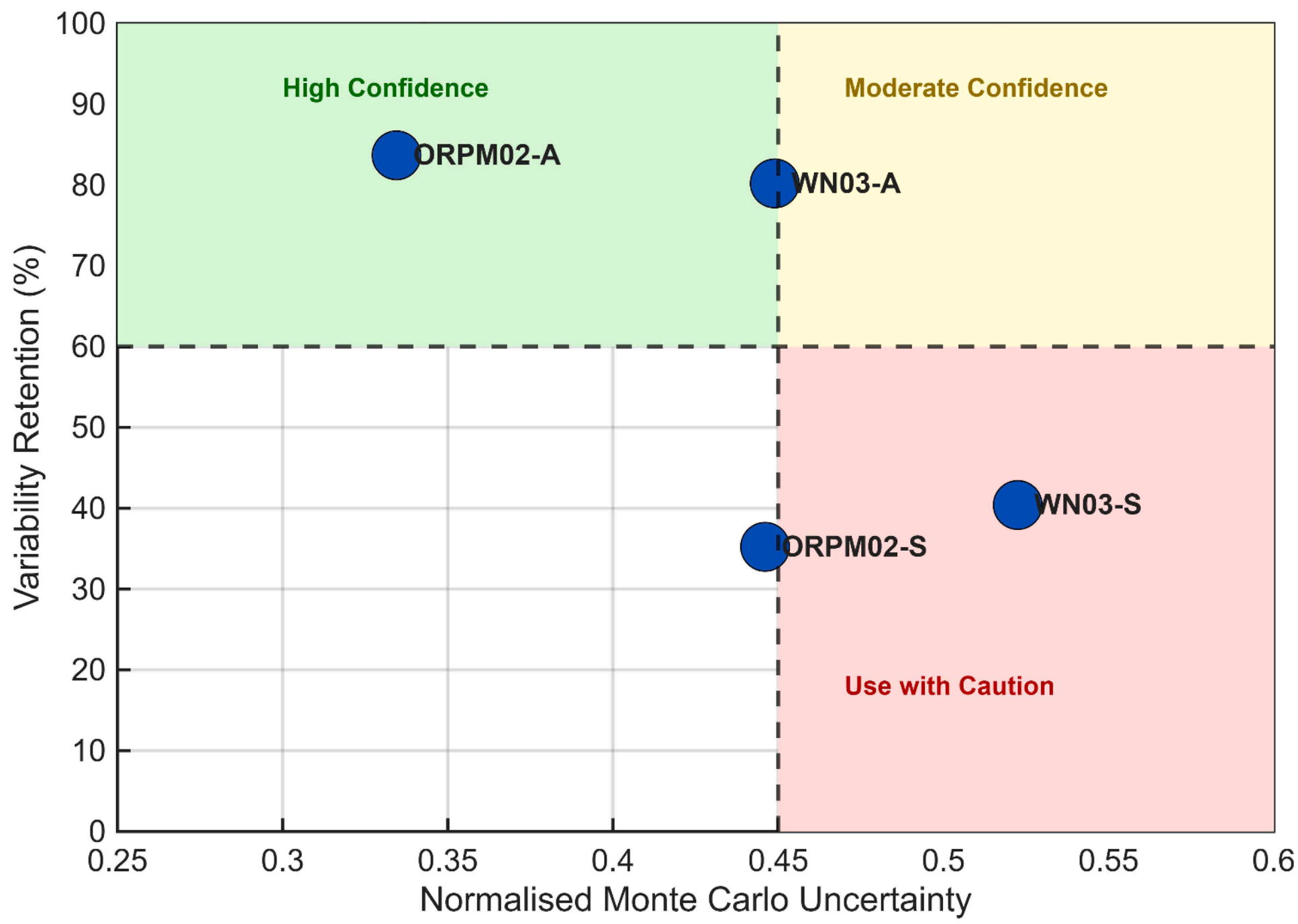

**Fig. 16.** Generator uncertainty versus variability of the measured responses.

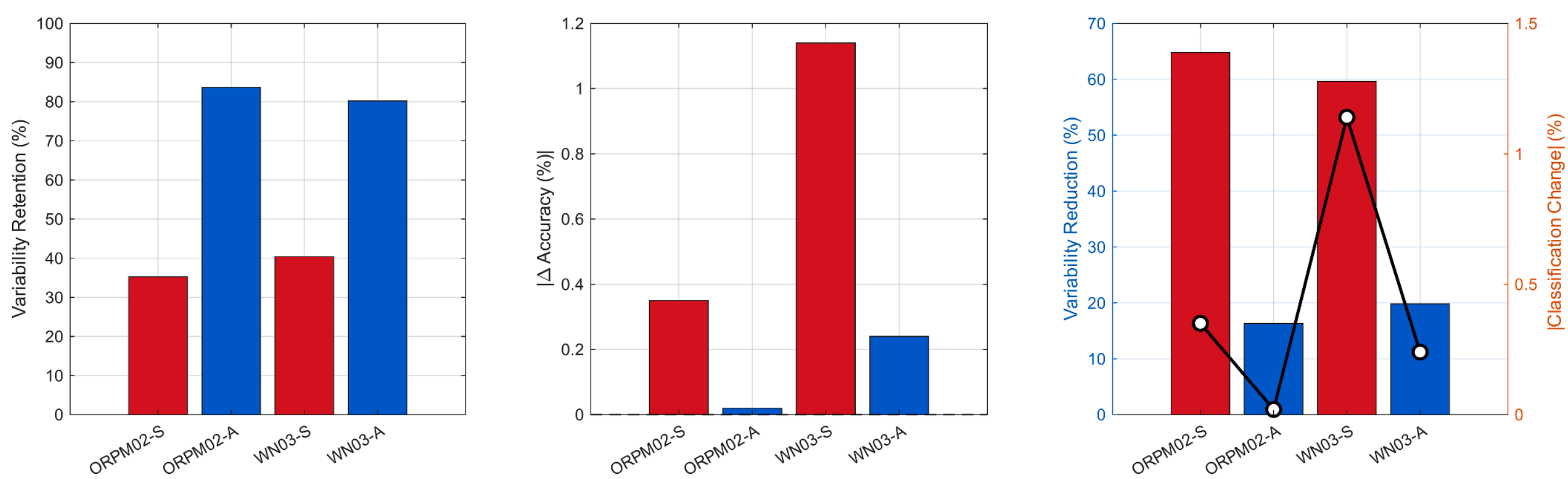

**Fig. 17.** The left plot shows the percentage of variability retained by the generated data relative to the experimental measurements, the middle plot shows the classification consistency, while the right plot compares variability reduction (100-Variability Retention) and classification change.

**Table 22**
Examined framework relative to conventional generative adversarial networks.

| Task | Conventional GAN | Examined Framework |
|---|---|---|
| Signal generation | Generation of synthetic structural responses. | Conditional generation of structural responses for healthy and damaged states. |
| Uncertainty information | Deterministic generation. | Monte Carlo dropout enables uncertainty-aware confidence assessment. |
| Validation strategy | Limited to visual inspection or classification performance. | Evaluation across time, frequency, classification, and uncertainty domains. |
| Classification assessment | Focused on demonstrating accuracy gains. | Quantifies both improvement and degradation. |
| Insight | Augmentation capability. | When augmentation is beneficial, and in low-separability structural monitoring. |

It should be emphasised that the assessment of generated data quality was not based solely on power spectral density comparisons. While PSD-based metrics provide insight into the reproduction of frequency-domain energy distributions, they do not by themselves demonstrate preservation of the complete structural response characteristics. For this reason, the generated signals were evaluated using multiple complementary analyses spanning time, frequency, variability, feature-space, classification, and uncertainty domains. Specifically, the assessment included waveform comparisons, RMS amplitude, coefficient of variation, PSD error metrics, spectral energy ratios, principal component analysis, support vector machine classification performance, and Monte Carlo dropout uncertainty quantification. The additional FRF characterisation presented in Fig. 2 further illustrates the underlying dynamic behaviour of the structure within the investigated frequency range. Consequently, the conclusions of this work are based on the collective agreement observed across multiple validation domains rather than on PSD similarity alone.

**Table 23**
Cross-dataset summary of the examined results obtained from all experimental configurations.

| Dataset | Sensor Type | Variability Reduction (%) | Accuracy Change (%) | Normalised Uncertainty |
|---|---|---|---|---|
| ORPM02-S | Strain | −64.79 | 0.35 | 0.446 |
| ORPM02-A | Acceleration | −16.32 | −0.02 | 0.327-0.342 |
| WN03-S | Strain | −59.60 | −1.14 | 0.521-0.524 |
| WN03-A | Acceleration | −19.82 | −0.24 | 0.434-0.464 |

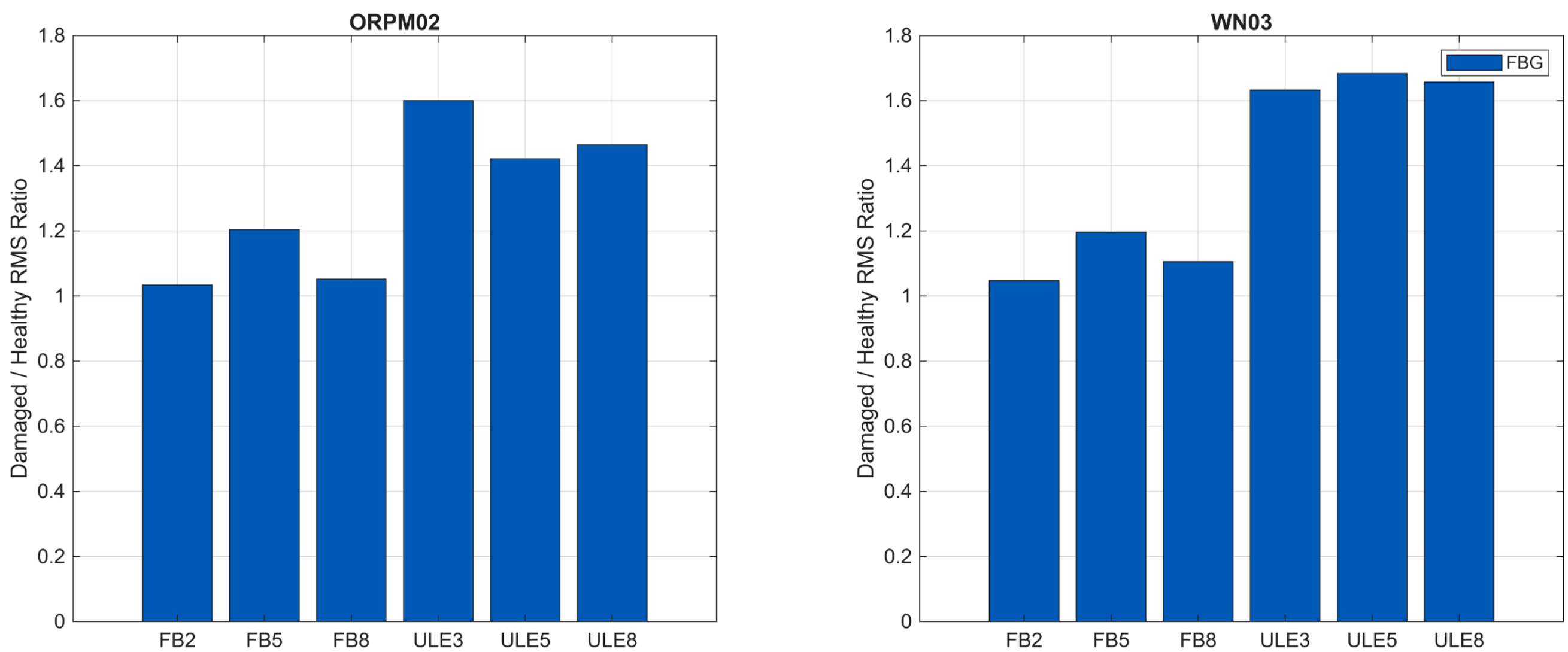

**Fig. 18.** Additional sensor validation using neighbouring FBG and accelerometer channels.

**Table 24**
Additional sensor validation using neighbouring FBG and accelerometer channels.

| Dataset | Sensor | Type | Healthy RMS | Damaged/Healthy Ratio |
|---|---|---|---|---|
| ORPM02 | PW_FB2 | FBG | 8.53 | 1.03 |
| ORPM02 | PW_FB5 | FBG | 7.27 | 1.20 |
| ORPM02 | PW_FB8 | FBG | 12.95 | 1.05 |
| ORPM02 | PW_ULE3 | ACC | 0.32 | 1.60 |
| ORPM02 | PW_ULE5 | ACC | 0.57 | 1.42 |
| ORPM02 | PW_ULE8 | ACC | 0.77 | 1.46 |
| WN03 | PW_FB2 | FBG | 7.85 | 1.05 |
| WN03 | PW_FB5 | FBG | 7.18 | 1.20 |
| WN03 | PW_FB8 | FBG | 12.13 | 1.11 |
| WN03 | PW_ULE3 | ACC | 0.42 | 1.63 |
| WN03 | PW_ULE5 | ACC | 0.71 | 1.68 |
| WN03 | PW_ULE8 | ACC | 1.14 | 1.66 |

Finally, Table 23 provides a direct comparison of the principal findings across all investigated datasets. The strain datasets exhibit substantially larger reductions in variability than the acceleration datasets, indicating greater difficulty in reproducing local strain behaviour. Classification performance is dataset dependent, explored for a range of 65% to 99% to ensure classification consistency for the generated signals. Finally, the uncertainty levels are consistently higher under white-noise excitation than under odd random-phase multisine excitation, suggesting a relationship between signal complexity and generator confidence. Further validation of the current observations using fully Bayesian approaches is suggested [55], as well as by using a larger number of datasets and sensor channels.

To further assess whether the selected sensor locations are representative of the structural response, additional accelerometer and fibre Bragg grating (FBG) measurements were examined. Table 24 and Fig. 18 show the damaged-to-healthy RMS amplification ratios for multiple sensor locations under both ORPM02 and WN03 excitation conditions. Although the absolute response amplitudes differ between sensors, the healthy-to-damaged amplification trends remain consistent across neighbouring measurement locations. This suggests that the observations reported in the main case studies are not unique to the originally selected sensors and are broadly representative of the local structural response behaviour within the investigated wing region.

## 7. Conclusion

This work examined the capability of an uncertainty-aware conditional generative adversarial network to synthesise structural vibration data for aerospace structural health monitoring. The framework was rigorously evaluated using full-scale experimental measurements from a BAE Systems Hawk T1A aircraft, including fibre Bragg grating strain and acceleration responses under multiple excitation and damage scenarios. The proposed approach enabled uncertainty-aware data augmentation with:

1. Class-conditional, long-range temporal and spectral characteristics learning from high-dimensional vibration data.
2. Integrated uncertainty quantification in generated signals and decision-making.
3. Identification of conditions under which data augmentation is consistent with the current damage classification performance.
4. Applicability across sensor modalities and excitation types.

Importantly, the examined framework provided a practical and effective generative component for vibration-based structural health monitoring, with potential generality extending beyond aerospace structures to other complex engineering systems, where representative damage data are limited and uncertainty-aware decision support is critical.

### CRediT authorship contribution statement

**Alvaro Peiro:** Writing – original draft, Visualization, Validation, Software, Methodology, Investigation, Formal analysis, Data curation, Conceptualization; **Marios Impraimakis:** Writing – review & editing, Supervision, Software, Methodology, Data curation, Conceptualization.

### Declaration of competing interest

The authors declare that they have no known competing financial interests or personal relationships that could have appeared to influence the work reported in this paper.

### Acknowledgements

The authors would like to gratefully acknowledge the University of Sheffield's Structural Dynamics Laboratory for Verification and Validation for providing the Hawk T1A aircraft.

### Data availability

The University of Sheffield's Structural Dynamics Laboratory for Verification and Validation provides the Hawk T1A aircraft data.